\documentclass[twocolumn]{svjour3}          % twocolumn
\smartqed  % flush right qed marks, e.g. at end of proof
\usepackage{natbib}
\usepackage{amsmath,amssymb,amsfonts}

\usepackage{textcomp}
\usepackage{xcolor}

\usepackage[misc]{ifsym}

\usepackage{times}
\usepackage{epsfig,subfig,graphicx,float}
\usepackage{booktabs,multirow,array} % For formal tables
\usepackage{url}

\usepackage{colortbl}
\newcommand{\gray}{\rowcolor[gray]{.9}}

\begin{document}

\title{Towards High Performance Human Keypoint Detection
\thanks{This work was supported by Australian Research Council Projects FL-170100117, DP-180103424, IH-180100002.}
}


\author{Jing Zhang \and Zhe Chen \and Dacheng Tao %etc.
}

\authorrunning{Jing Zhang, et al.} % if too long for running head

\institute{
\Letter~Zhe Chen (zhe.chen1@sydney.edu.au)\\
\Letter~Dacheng Tao (dacheng.tao@sydney.edu.au)\\
Authors are with School of Computer Science, Faculty of Engineering, The University of Sydney, Darlington, NSW 2008, Australia
}

\date{Received: date / Accepted: date}

\maketitle

\begin{abstract}
Human keypoint detection from a single image is very challenging due to occlusion, blur, illumination, and scale variance. In this paper, we address this problem from three aspects by devising an efficient network structure, proposing three effective training strategies, and exploiting four useful postprocessing techniques. First, we find that context information plays an important role in reasoning human body configuration and invisible keypoints. Inspired by this, we propose a cascaded context mixer (CCM), which efficiently integrates spatial and channel context information and progressively refines them. Then, to maximize CCM's representation capability, we develop a hard-negative person detection mining strategy and a joint-training strategy by exploiting abundant unlabeled data. It enables CCM to learn discriminative features from massive diverse poses. Third, we present several sub-pixel refinement techniques for postprocessing keypoint predictions to improve detection accuracy. Extensive experiments on the MS COCO keypoint detection benchmark demonstrate the superiority of the proposed method over representative state-of-the-art (SOTA) methods. Our single model achieves comparable performance with the winner of the 2018 COCO Keypoint Detection Challenge. The final ensemble model sets a new SOTA on this benchmark. The source code will be released at \url{https://github.com/chaimi2013/CCM}.

\keywords{Human Pose Estimation \and Deep Nerual Networks \and Sub-pixel Refinement \and Context}

\end{abstract}

\section{Introduction}
\label{sec:introduction}

%[htbp]
\begin{figure}[t]
\centering
\subfloat[]{\label{fig:coco_stats_train_occlusions}%%
\includegraphics[width=0.49\linewidth]{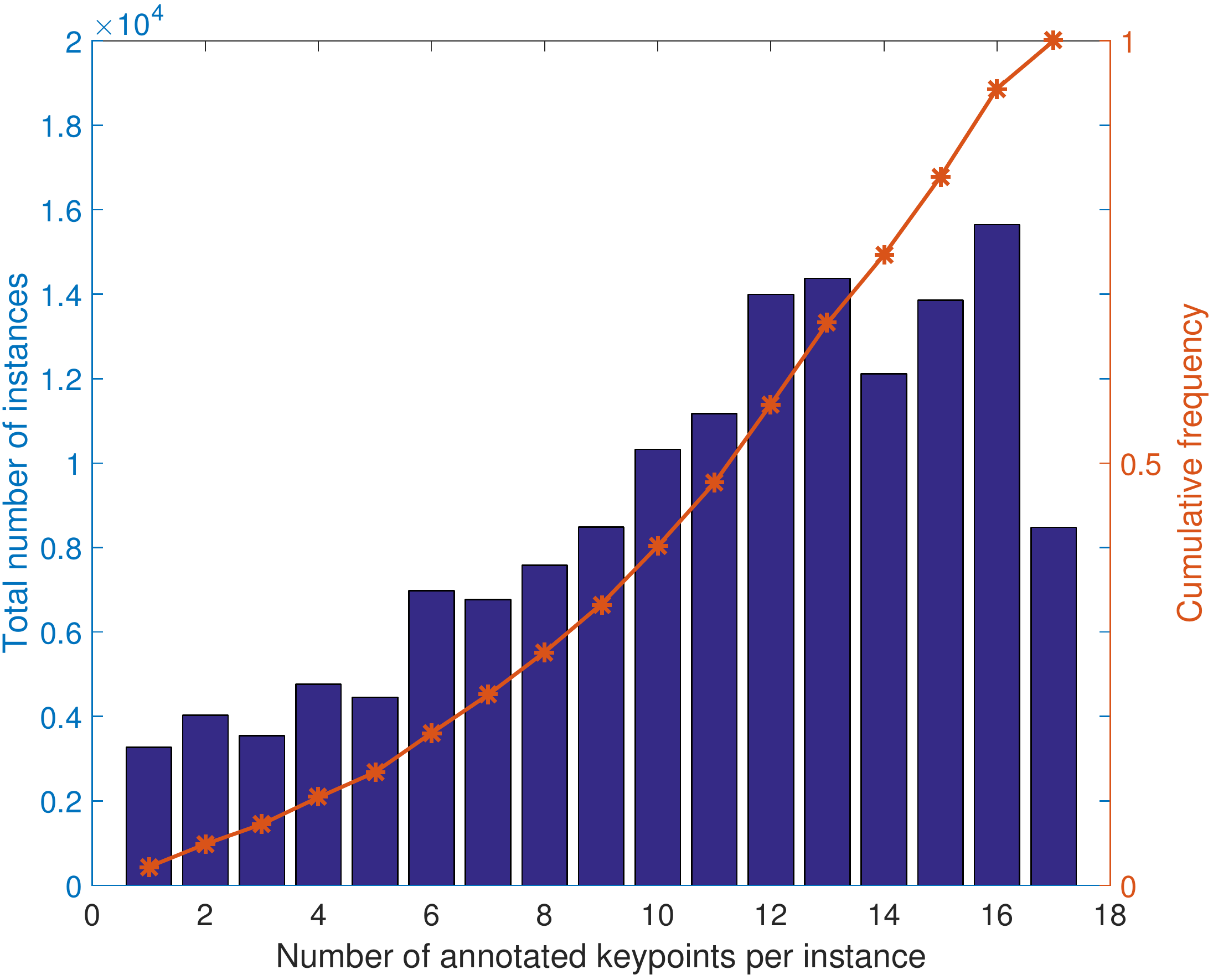}}
\vspace{0.00001\linewidth}
\subfloat[]{\label{fig:coco_statas_train_vis_invis}%%
\includegraphics[width=0.49\linewidth]{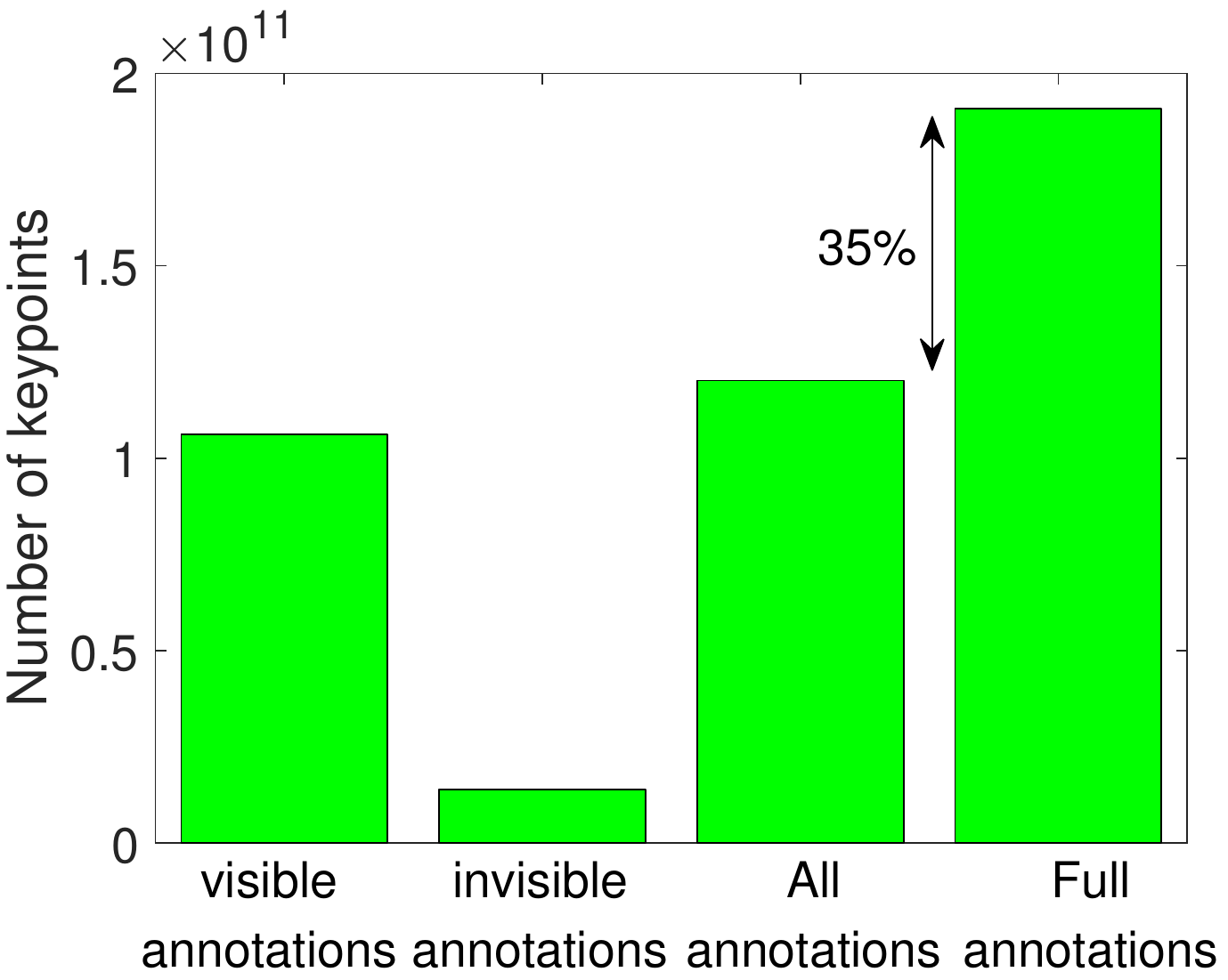}}
\vspace{0.00001\linewidth}
\subfloat[]{\label{fig:coco_stats_imgs}%%
\includegraphics[width=1\linewidth]{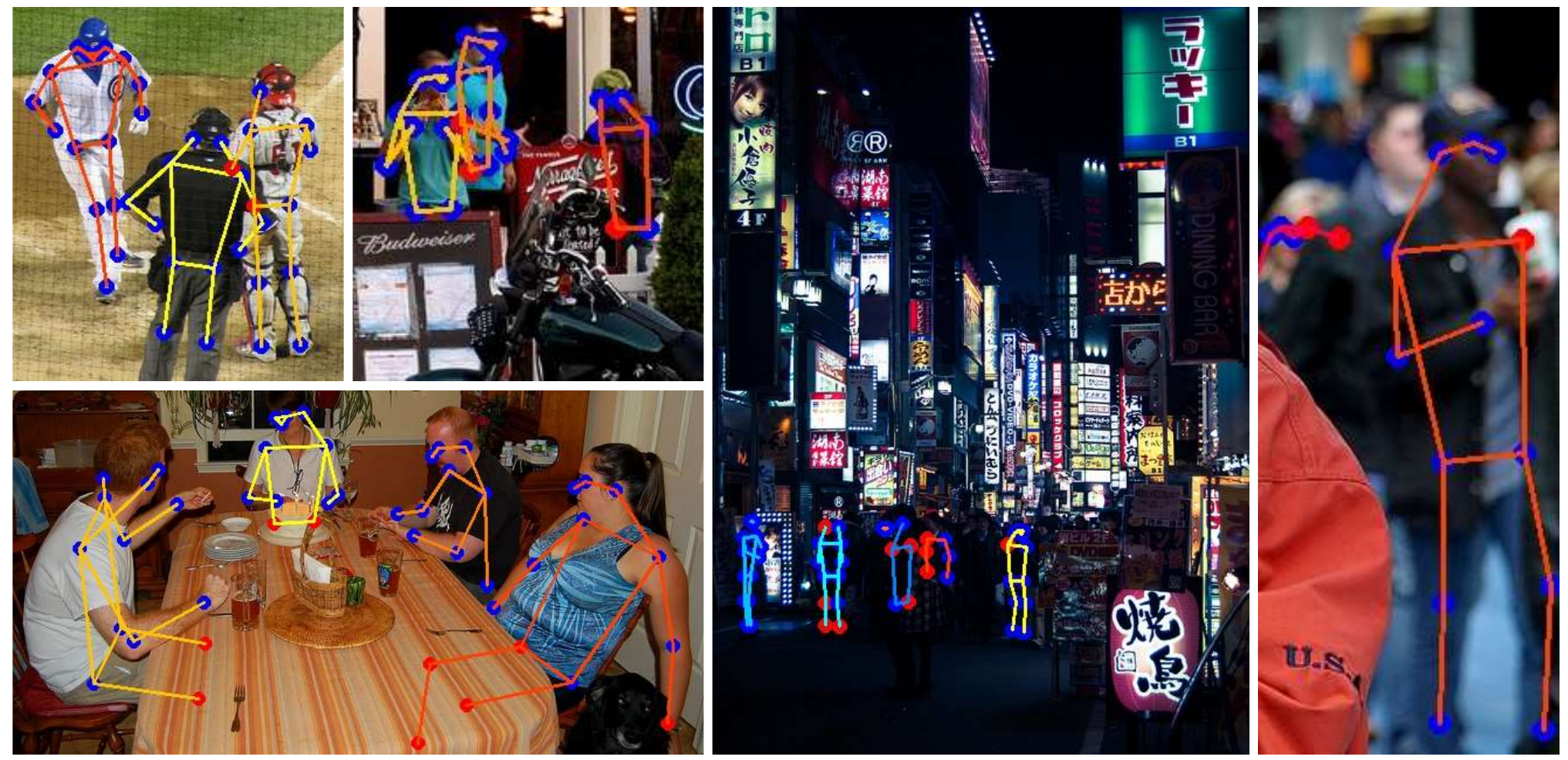}}

\caption{(a)-(b) Statistics of the annotated keypoints in the MS COCO training dataset \citep{lin2014microsoft}. (c) Some examples from the MS COCO dataset, where occluded, under-exposed, and blurry person instances are very common. Blue and red dots denote the annotated visible and invisible keypoints, respectively. }
\label{fig:coco_stats}
\end{figure}

Human keypoint detection is also known as human pose estimation (HPE) refers to detecting human body keypoint location and recognizing their categories for each person instance from a given image. It is very useful in many downstream applications such as activity recognition \citep{ni2017learning, baradel2018glimpse, liu2018skeleton}, human-robot interaction \citep{mazhar2018towards,zhang2020empowering}, and video surveillance \citep{hattori2018synthesizing, varadarajan2018joint}. However, HPE is very challenging even for human annotators. For example, almost 50\% of instances\footnote{The instances with at least one annotated keypoint are counted.} in the COCO training dataset \citep{lin2014microsoft} have at least six unannotated keypoints (Figure~\ref{fig:coco_stats}(a)) and 35\% keypoints are unannotated (Figure~\ref{fig:coco_stats}(b)) due to various factors including occlusion, truncation, under-exposed imaging, blurry appearance and low-resolution of person instances.  Some examples from the MS COCO dataset are shown in Figure~\ref{fig:coco_stats}.

Prior methods have made significant progress in this area with the success of deep convolutional neural networks (DCNN) \citep{toshev2014deeppose}, which can be categorized into two groups: top-down methods and bottom-up methods. Top-down methods are composed of two phases, i.e., person detection and keypoint detection, while bottom-up methods directly detect all keypoints from the image and associate them with corresponding person instances. Although bottom-up methods \citep{cao2017realtime} are usually fast, top-down methods still dominate the leaderboard of public benchmark datasets like MS COCO due to their high accuracy. Using heatmaps to represent keypoint locations and fully CNNs with an encoder-decoder structure to learn features has gained prominence in recent studies \citep{newell2016stacked,huang2017coarse}, since spatial correspondence can be preserved between feature maps and heatmaps. Recently, to learn strong feature representations at multiple resolutions, pyramid-based networks have been proposed \citep{yang2017learning,chen2018cascaded}. Although these methods improve detection accuracy by learning multi-scale features, there are still some issues to be addressed.

First, detecting invisible keypoints is more difficult than visible ones due to ambiguous appearance and inconsistent context bodies. How to effectively model multi-source context information to infer the hard keypoints is still under-explored. Second, external datasets such as the AI Challenger dataset\footnote{\url{https://challenger.ai/competition/keypoint/}}, have been used to learn more discriminative feature representations \citep{xiao2018simple,sun2019deep}. However, they may have different annotation formats with the target set, for example, 17 keypoints for the MS COCO dataset and 14 keypoints for the AI Challenger dataset, or even have no annotations like the MS COCO unlabeled dataset. How to effectively leverage these external datasets to learn human pose configuration and discriminative feature representations for recognizing diverse poses remains challenging. Third, the ground truth keypoint location is annotated in pixel (or sub-pixel) in the high-resolution image plane, while the regression target is usually in the low-resolution heatmap, e.g., 1/4 size of the input image. This scale mismatch of representation will degrade the keypoint detection performance. To address this issue, sub-pixel representation or post-processing techniques \citep{chen2018cascaded,zhang2020distribution} have been proposed. Nonetheless, a systematic study of post-processing techniques is still absent and worth further discovering.

In this paper, we address these issues to improve human keypoint detection by devising an efficient network structure, proposing three effective training strategies, and exploiting four useful postprocessing techniques. First, inspired by the keypoint detection process carried out by humans, where the ``context'' contributes to the perception and inference process, we advance the research by studying the role of context information for human keypoint detection. Specifically, we adopt an encoder-decoder network structure and propose a novel cascaded context mixer (CCM) in the decoder. It can efficiently integrate both spatial and channel context information and progressively refine them. Then, we propose a joint training strategy and a knowledge-distilling approach to exploit abundant unlabeled data. Besides, we also propose a hard-negative person detection mining strategy to migrate the inconsistency of person instances between training and testing. These strategies endow the detection network with the capability of learning discriminative features. Third, we present and comprehensively evaluate four sub-pixel refinement techniques for postprocessing keypoint predictions. Extensive experiments on the MS COCO keypoint detection benchmark validate the effectiveness of the proposed CCM model, the training strategies, and the sub-pixel techniques. Our single model achieves comparable performance with the winner of the 2018 COCO Keypoint Detection Challenge. The final ensemble model sets a new SOTA on this benchmark\footnote{A video demo can be found in \url{https://github.com/chaimi2013/CCM/video}}.

The contributions of this work can be summarized as follows:

$\bullet$ We devise an effective cascaded context mixer in the decoder which can learn both spatial and channel context information to infer the human body and hard keypoints.

$\bullet$ We propose several efficient training strategies to guide our model to deal with false-positive person detections and learn discriminative features from diverse poses.

$\bullet$ We present some sub-pixel refinement techniques to enhance location accuracy and comprehensively evaluate their performance and complementarity.

\section{Related work}
\label{sec:relatedwork}

\begin{table*}[htbp]
  \centering
  \caption{A summary of the human keypoint detection methods based on DCNN.}
    \begin{tabular}{llccccc}
    \toprule
   Category  & Method & Backbone  & Decoder  & Extra Data & Post-processing & Performance \\
    \midrule
     \multicolumn{1}{c}{\multirow{3}[0]{*}{Bottom-up}} & \cite{pishchulin2016deepcut} & VGG & - & - & offset regression & 82.4PCK$_{h}$@MPII\\
    \multicolumn{1}{c}{} & \cite{cao2017realtime} & VGG-19 &  multi-stage CNN & - & - & 61.8AP@COCO\\
    \multicolumn{1}{c}{} & \cite{newell2017associative} & Hourglass & - & - & multiscale average & 65.5AP@COCO\\
    \midrule
    \midrule
    \multicolumn{1}{c}{\multirow{9}[0]{*}{Top-down}} & \cite{he2017mask} & ResNet-50-FPN & conv+deconv & - & offset regression & 63.1AP@COCO\\
    \multicolumn{1}{c}{} & \cite{fang2017rmpe} & STN+Hourglass & - & - &  parametric NMS & 63.3AP@COCO\\
    \multicolumn{1}{c}{} & \cite{papandreou2018personlab} & ResNet-101 & 1x1 conv & \checkmark & offset regression & 68.5AP@COCO\\
    \multicolumn{1}{c}{} & \cite{chen2018cascaded} & ResNet-Inception & GlobalNet & - & flip/GF & 72.1AP@COCO\\
    \multicolumn{1}{c}{} & \cite{xiao2018simple} & ResNet-152 & deconv & - & flip/sub-pixel shift & 76.5AP@COCO\\
    \multicolumn{1}{c}{} & \cite{sun2019deep} & HRNet-w48 & 1x1 conv & - & flip/sub-pixel shift & 77.0AP@COCO\\
    \multicolumn{1}{c}{} & \cite{li2019rethinking} & 4 x ResNet-50 & GlobalNet & \checkmark & flip/GF/sub-pixel shift & 78.1AP@COCO\\
    \multicolumn{1}{c}{} & \multicolumn{1}{l}{\multirow{2}[0]{*}{\textbf{Our CCM}}} & ResNet-152 & \multicolumn{1}{c}{\multirow{2}[0]{*}{CCM}} & \multicolumn{1}{c}{\multirow{2}[0]{*}{\checkmark}} & \multicolumn{1}{c}{\multirow{2}[0]{*}{sub-pixel refinement}} & \multicolumn{1}{c}{\multirow{2}[0]{*}{78.9AP@COCO}}\\
    \multicolumn{1}{c}{} & \multicolumn{1}{c}{} & HRNet-w48 & \multicolumn{1}{c}{} &\multicolumn{1}{c}{} & \multicolumn{1}{c}{} & \multicolumn{1}{c}{}\\
    \bottomrule
    \end{tabular}%
  \label{tab:methodsummary}%
\end{table*}%

HPE methods can be grouped into 2D pose estimation \citep{rogez2012fast,toshev2014deeppose, newell2016stacked, fang2017rmpe, huang2017coarse, cao2017realtime, yang2017learning, xiao2018simple, sun2018integral, chen2018cascaded, sun2019deep, zhang2019human, li2019rethinking, girdhar2018detect} and 3D pose estimation \citep{rogez2012fast,pavlakos2018learning, pavlakos2018ordinal, rhodin2018unsupervised, hossain2018exploiting, yang20183d} according to the dimension of the coordinates of the keypoint locations. In this paper, we focus on 2D pose estimation, specifically, the multi-person pose estimation (MPPE) problem, which is more challenging than the single-person pose estimation (SPPE) problem. MPPE approaches can be further divided into bottom-up and top-down approaches. A summary of these approaches is presented in Table~\ref{tab:methodsummary}, where we outline their features according to the network structure, whether using extra data or not, and the postprocessing techniques used to improve detection accuracy. The details are presented in the following part.

Bottom-up approaches first detect all human keypoints and then associate them with each detected person instance \citep{cao2017realtime,newell2017associative,pishchulin2016deepcut}. This approach is usually faster than top-down approaches, but the assembly step can become intractable when person instances are ambiguous due to occlusions, blur, etc., degrading accuracy compared to top-down approaches. Top-down approaches first detect all person instances in the image and then apply SPPE on each detected person. Benefiting from recent progress in DCNN-based object detection \citep{ren2015faster, ouyang2016learning, he2017mask, lin2017feature, liu2020deep,chen2020recursive}, person detectors have achieved promising detection accuracy. Further, different neural networks have been proposed to learn strong feature representations based on multi-scale feature fusion and multi-level supervisions, \emph{e.g.}, the Pyramid Residual Module in \citep{yang2017learning}, the Cascaded Pyramid Network in \citep{chen2018cascaded}, the simple baseline model in \citep{xiao2018simple}, and the High-resolution Net in \citep{sun2019deep}, which detect keypoint locations with high accuracy. Our proposed method follows the top-down scheme and has an encoder-decoder structure. However, in contrast to the above methods, we study the role of context information for human keypoint detection by devising a cascaded context mixer module in the decoder to sequentially capture both spatial and channel context information.

Deep neural models benefit from a large scale of training data and efficient training strategies. Trained on more examples with diverse poses, the keypoint detection model can learn to infer the occluded or blurry keypoints from similar poses \citep{xiao2018simple,li2019rethinking}. For example, all the top entries of the COCO keypoint detection leaderboard\footnote{\url{http://cocodataset.org/index.htm#keypoints-leaderboard}} leverage external data such as the AI Challenger human keypoint detection dataset. In this paper, we also validate the benefit of using external data. However, instead of using transfer learning, we propose a more effective joint training strategy to harvest external data with heterogeneous labels. Further, we make use of unlabeled data, e.g., the unlabeled MS COCO dataset, referring to the knowledge distilling idea, where the pseudo-labels are generated by a teacher model. Besides, few of the above approaches deal with the mismatch problem of person detections during the training phase and testing phase, i.e., training with ground truth person instances while testing with detected ones, which may be false positives. In this paper, we propose an effective hard-negative person detection mining strategy in the training phase, adapting the model to predict no keypoints for those false person instances.

Dominant methods adopt a heatmap to represent the keypoint location where a Gaussian density map is placed on the corresponding pixel. However, the heatmap is usually in low-resolution compared with the input, which has a side effect on the location accuracy. Increasing the heatmap resolution means to decode high-resolution features, which may incur extra computational cost and model complexity. Instead, prior methods adopt computationally efficient postprocessing techniques to refine the predictions, for example, shifting the detected location by 0.25 pixels according to the local gradient directions \citep{chen2018cascaded,xiao2018simple,sun2019deep}. In contrast to them, we present a sub-pixel refinement technique using the second-order approximation, it turns out to be more effective than the above one and boost the detection accuracy by a large margin. Besides, we comprehensively study the sub-pixel refinement techniques for postprocessing including the second-order approximation (SOA), the Soft Non-Maximum Suppression (Soft-NMS), the sub-pixel shift of flipped heatmaps (SSP), and the Gaussian filtering on heatmaps (GF). Experiment results validate that these techniques can significantly boost the keypoint detection performance. Moreover, they are complementary to each other, and so the combination of them will boost the performance further.

\section{Cascaded Context Mixer for Human Keypoint Detection}
\label{sec:method}
In this part, we propose a novel human keypoint detection model based on a cascaded context mixer module to explicitly and simultaneously model spatial and channel context information. To further exploit the representation capacity of the model, we propose three efficient training strategies including a hard-negative person detection mining strategy to migrate the mismatch between training and testing, a joint-training strategy to use abundant unlabeled samples by knowledge distilling, and a joint-training strategy to exploit external data with heterogeneous labels. To improve the detection accuracy, we also present four postprocessing techniques to refine predictions at the sub-pixel level.

\subsection{Motivation}
\label{subsec:motivation}
\begin{figure}[t]
\centering
\subfloat[]{\label{fig:motivation_occlusion}%%
\includegraphics[width=1\linewidth]{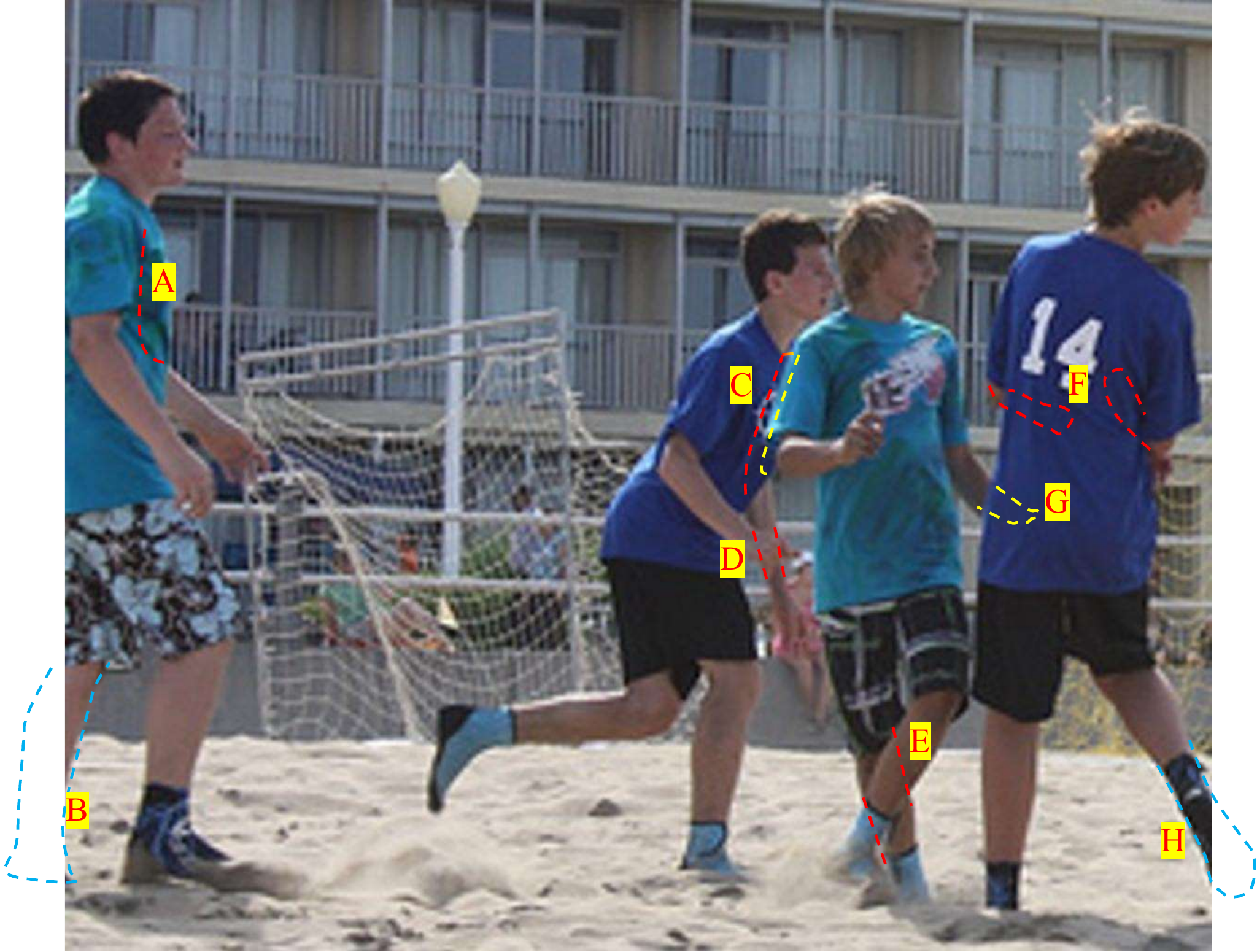}}
\vspace{0.0001\linewidth}
\subfloat[]{\label{fig:motivation_stage}%%
\includegraphics[width=1\linewidth]{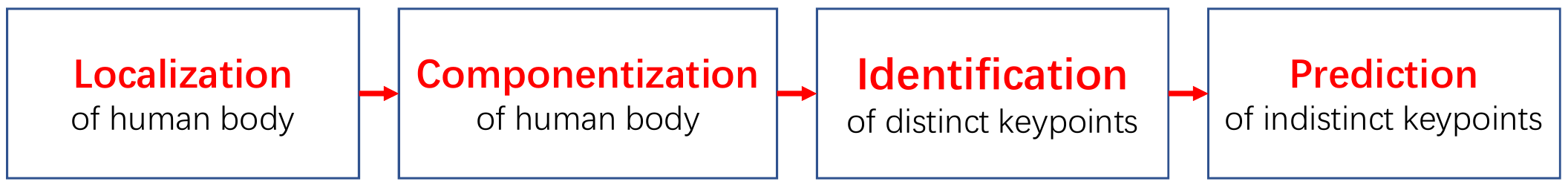}}

\caption{(a) Ubiquitous occlusions in an image from the MS COCO dataset. (b) Illustration of the keypoint detection process carried out by humans when faced with occlusions.}
\label{fig:motivation}
\end{figure}

Since occlusions are ubiquitous in and between human bodies, we take it as an example to show how humans carry out the detection process when dealing with the aforementioned hard cases. As shown in Figure~\ref{fig:motivation}(a), occlusions include self-occlusion (A, C, D, E, F), occlusion by others (C, G), and truncation (B, H). Note that all the faces are self-occluded in Figure~\ref{fig:motivation}(a), \emph{i.e.}, half of each face is invisible. Humans can easily recognize a complete object and its boundaries, even if it is occluded. According to Gestalt psychology in visual perception, we can group fragmented contours under the law of closure \citep{wagemans2012century}, e.g., D and E. We have also seen numerous human body positions and acquired the common sense that a body consists of symmetric legs, hands, and face and that different body parts move and form different poses. Therefore, we can easily infer the occluded parts like A, B, C, G, and H. For the more difficult case F, we may infer the invisible arms by judging the boy's intention and rehearsing the same action psychologically.

%[htbp]

The Recognition-by-Components (RBC) theory tells us that humans quickly recognize objects even under occlusions by characterizing the object's components \citep{biederman1987recognition}. One possible path of the keypoint detection process carried out by humans could be divided into four main stages as shown in Figure~\ref{fig:motivation}(b). First, we recognize and locate a human body (The top-down approaches follow this paradigm \citep{chen2018cascaded,xiao2018simple,sun2019deep}). Then, we recognize each body part to further locate the keypoints belonging to it (Some graph-based models complete this step explicitly \citep{felzenszwalb2008discriminatively, holt2011putting, wang2013beyond, yang2013articulated}). Here, we can easily identify some distinct and visible keypoints. Finally, we infer the remaining keypoints which may be invisible or ambiguous. How do we accomplish this? By reviewing the inference process and the occlusion cases in Figure~\ref{fig:motivation}(a), we think that the ``\emph{context}'' plays an important role when we associate separate body parts into a whole or infer an invisible keypoint \citep{chen2020recursive,ma2020auto}. Another key factor may be that we have \emph{a priori} knowledge of human body configurations in all possible poses. The context tells us about the surrounding visible body parts, and the \emph{a priori} knowledge helps us to determine what the category and location of the invisible part should be. This motivates us to design a context-aware model that can efficiently learn useful feature representation from diverse poses.

\begin{figure}[t]
\centering
\includegraphics[width=1\linewidth]{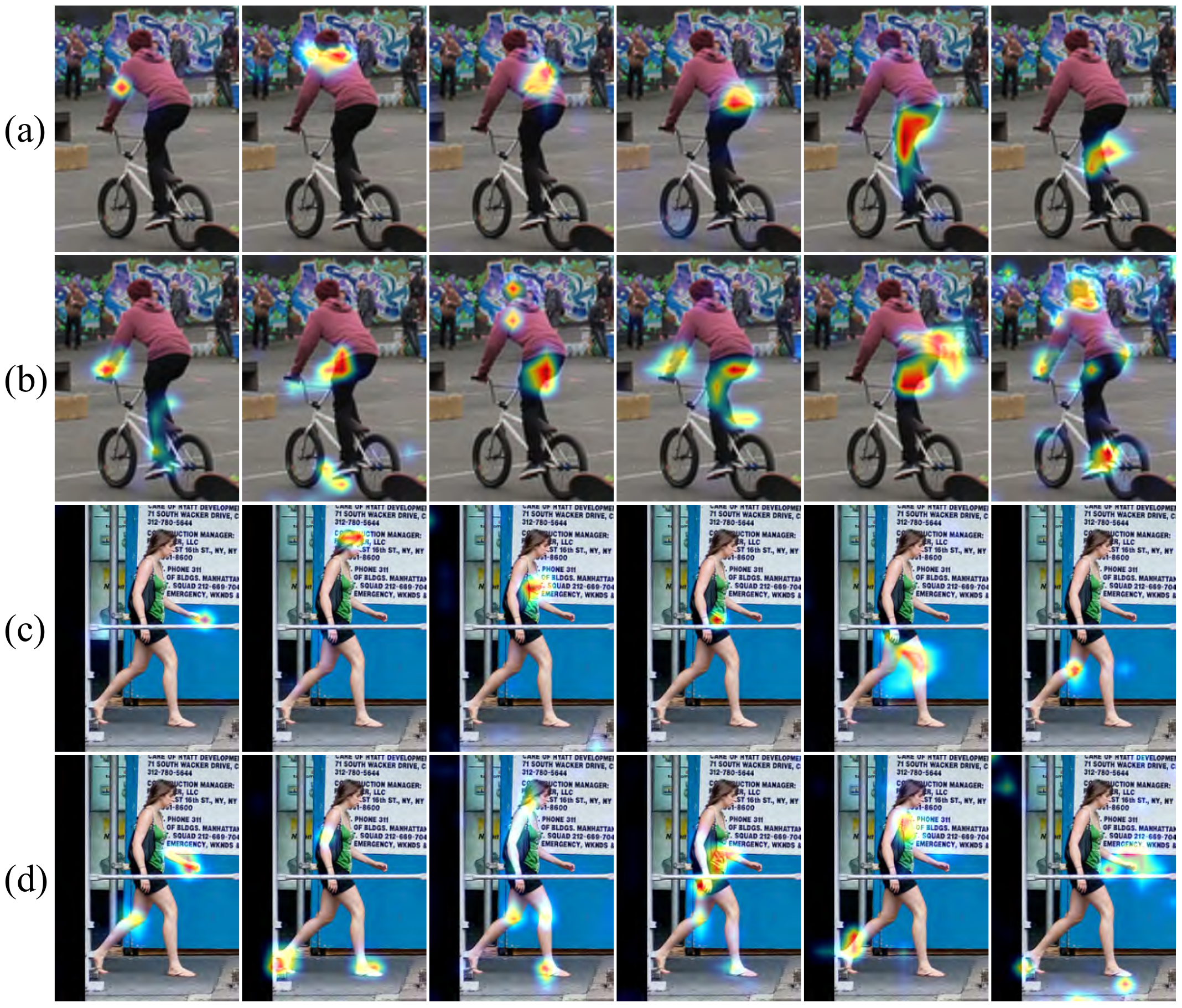}
\caption{Visualization of the feature maps learned by the ResNet-50 encoder on two test images as shown in (a)-(b) and (c)-(d), respectively.}
\label{fig:feat_vis_res4}
\end{figure}

\begin{figure*}[t]
\centering
\includegraphics[width=1\linewidth]{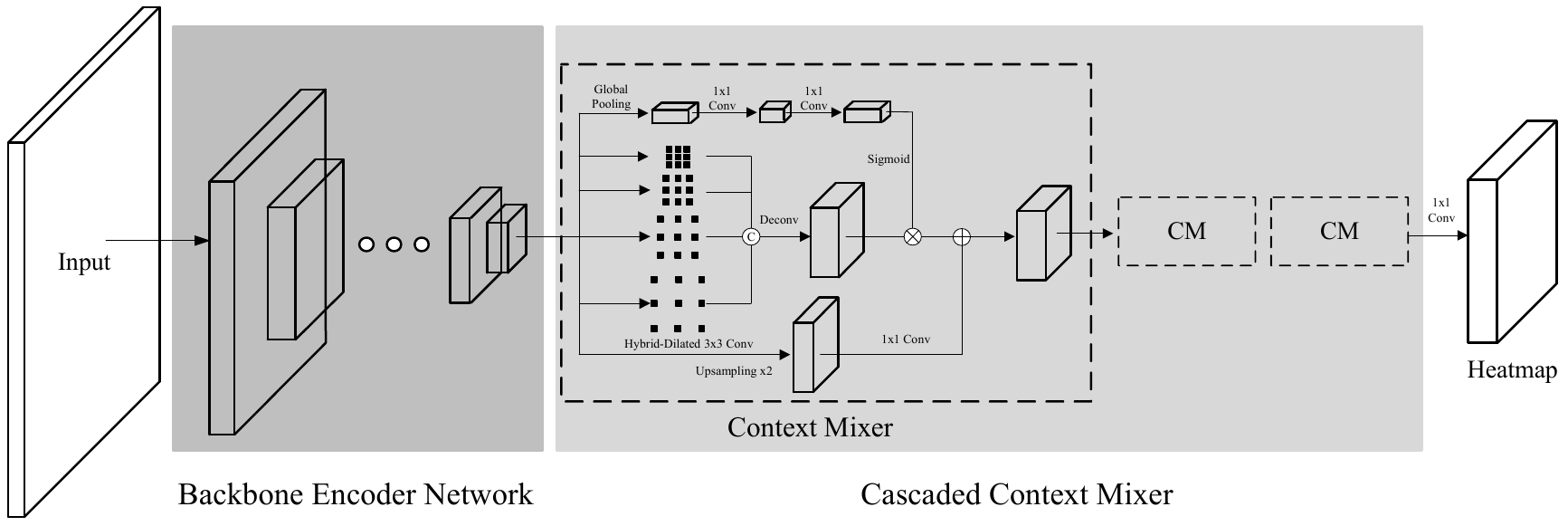}
\caption{The diagram of the proposed human keypoint detection model based on cascaded context mixer (CCM).}
\label{fig:CCM}
\end{figure*}

\subsection{Cascaded Context Mixer-based Decoder}
\label{subsec:cam}
In this paper, we tackle the multi-person pose estimation problem by following a topdown scheme. First, a human detector is used to detect the bounding box for each person instance. Then, the proposed CCM model detects keypoints for each person instance. After aggregating the detections using Object Keypoint Similarity (OKS)-based Non-Maximum Suppression (NMS), we obtain the final pose estimation. As shown in Figure~\ref{fig:CCM}, the CCM model has an encoder-decoder structure where we use ResNet \citep{he2016deep} or HRNet \citep{sun2019deep} as the backbone encoder network. The decoder consists of three Context Mixer (CM) modules in a cascaded manner. The details of CM are presented as follows.

\subsubsection{Context Mixer}
\label{subsubsec:CM}
One the one hand, as we know that DCNN is able to learn distributed semantics at each feature channel, which leads to powerful representation ability to recognize a vast number of categories. In Figure~\ref{fig:feat_vis_res4}, we visualized some feature channels from the ResNet-50 encoder. As can be seen, the encoder learned to activate one or multiple explicit body parts. On the other hand, as discussed in Section~\ref{subsec:motivation}, context information could be useful to infer invisible keypoints and help to learn human body configuration. Thereby, we exploit two types of context information in this paper, i.e., global context and local context.

First, since different feature channels correspond to certain body parts (Figure~\ref{fig:feat_vis_res4}(a) and Figure~\ref{fig:feat_vis_res4}(c)), the relationship between them can be modeled to further enhance the feature representation. To this end, we use a channel attention module to encode the relationship between different channels, which inherits the idea of squeeze-and-excitation (SE) network \citep{hu2018squeeze}. The attention vector is used to reweigh features in a channel-wise manner, which is a global operation from the perspective of spatial dimension. In this way, we can extract the global context to enhance the feature representation. Second, since some of the feature responses at different body parts have strong correlations (Figure~\ref{fig:feat_vis_res4}(b) and Figure~\ref{fig:feat_vis_res4}(d)), we can encode the spatial relationship by using convolution layers with large receptive fields such that they can cover different body parts. To this end, we leverage hybrid-dilated convolutions to capture multi-scale spatial context information within different receptive fields, which inherits the idea of atrous spatial pyramid pooling (ASPP) \citep{chen2018deeplab}. These two types of context information collaborate with each other to learn discriminative feature representations. Besides, we also use a residual branch to reuse the learned features from the previous stage. Thereby, there are three branches in CM as shown in the middle part of Figure~\ref{fig:CCM}. We present the details of each branch as follows.

In the residual branch of the $k^{th}$ CM, feature maps $f_{k-1}^{CM}$ from the previous stage are first up-sampled two times before being fed into a $1 \times 1$ convolutional layer to output feature maps $f_k^{RES}$ of size $H_k \times W_k \times C_k$, i.e., 
\begin{equation}
f_k^{RES} = \phi_k^{RES} \left( f_{k-1}^{CM} \uparrow_2 \right),
\label{eq:feat_res}
\end{equation}
where $\uparrow_2$ denotes the up-sampling operation, $\phi_k^{RES}\left(\cdot\right)$ denotes the function learned by the convolutional layer.

In the SE branch of the $k^{th}$ CM, the feature maps $f_{k-1}^{CM}$ first go through a global pooling layer. Then, the obtained feature vector is fed into a bottle-neck layer with $1 \times 1$ convolutions. The feature dimension is reduced to $1 \times 1 \times {{{C_k}} \mathord{\left/
 {\vphantom {{{C_k}} 4}} \right.
 \kern-\nulldelimiterspace} 4}$. Then, it is fed into a subsequent $1 \times 1$ convolutional layer to increase the feature dimension to $1 \times 1 \times C_{k}$. A sigmoid function is used to squeeze the feature vector $f_k^{SE}$ into the range $\left[ {0,1} \right]$, i.e.,
\begin{equation}
\alpha_k^{SE} = \sigma \left( \phi_k^{SE} \left( GP \left( f_{k-1}^{CM} \right) \right) \right),
\label{eq:feat_se}
\end{equation}
where $GP\left(\cdot\right)$ denotes the global pooling operation, $\phi_k^{SE}\left(\cdot\right)$ denotes the function learned by those two $1 \times 1$ convolutional layers, and $\sigma\left(\cdot\right)$ denotes the sigmoid activation function.

In the HDC branch of the $k^{th}$ CM, the feature maps $f_{k-1}^{CM}$ go through four $3 \times 3$ convolutional layers with different dilated rates, \emph{i.e.}, 1, 2, 3, and 4. Each convolutional layer has ${{{C_k}} \mathord{\left/
 {\vphantom {{{C_k}} 4}} \right.
 \kern-\nulldelimiterspace} 4}$ kernels. These feature maps are then concatenated and fed into a deconvolutional layer of stride 2. The output feature maps $f_k^{HDC}$
are of size $H_k \times W_k \times C_k$, i.e.,
\begin{equation}
f_k^{HDC} = \phi_k^{HDC} \left( \left [ \phi_k^{d1} \left( f_{k-1}^{CM} \right);\dots;\phi_k^{d4} \left( f_{k-1}^{CM} \right) \right ] \right),
\label{eq:feat_hdc}
\end{equation}
where $\phi_k^{d1}\left(\cdot\right) \sim \phi_k^{d4}\left(\cdot\right)$ denote functions learned by the dilated convolutional layers, $\phi_k^{HDC}\left(\cdot\right)$ denotes the function learned by the deconvolutional layer, and $\left [ ; \right ]$ denotes the concatenation operation.

Then, the output of the $k^{th}$ CM is calculated as:
\begin{align}\nonumber
f_k^{CM} &= f_k^{HDC} \odot \alpha_k^{SE} + f_k^{RES}\\
    &\triangleq \phi_k^{CM} \left( f_{k-1}^{CM} \right),
\label{eq:feat_cam}
\end{align}
where $\odot$ denotes the channel-wise multiplication and $\phi_k^{CM} \left(\cdot\right)$ denotes the mapping function learned by the $k^{th}$ CM. Batch normalization \citep{ioffe2015batch} is used after each convolutional layer and deconvolutional layer. ReLU is used after the first convolutional layer in the SE branch, all the convolutional layers in the HDC branch, and the output of each CM module \citep{krizhevsky2012imagenet}.

\subsubsection{Cascaded Context Mixer}
\label{subsubsec:CCM}
To capture the context information at multiple resolutions and learn multi-scale feature representation, we stack $K$ CMs sequentially to decode the features from the encoder step by step and increase their resolutions accordingly. Mathematically, it can be written as:
\begin{equation}
f^{ENC} = \phi^{ENC}\left( img \right),
\label{eq:encoder}
\end{equation}
\begin{align}\nonumber
 f^{DEC} &= \phi_K^{CM}\left( {\dots \left( {\phi_1^{CM}\left( f^{ENC} \right)} \right)} \right) \\
    &\triangleq \phi^{DEC} \left( f^{ENC} \right)
  \label{eq:decoder}
\end{align}
where $img$ represents the input image, $\phi^{ENC}\left(  \cdot  \right)$ denotes the function learned by the encoder, $f^{ENC}$ is the encoded feature, $\phi^{DEC}\left(  \cdot  \right)$ denotes the function learned by the decoder, $f^{DEC}$ is the decoded feature. Note that we denote $f_{0}^{CM} \triangleq f^{ENC}$ for consistency.

The decoded feature $f^{DEC}$ is fed into a final $1 \times 1$ convolutional layer to predict the target heatmaps:
\begin{equation}
h = \phi^{PRE} \left( f^{DEC} \right),
\label{eq:predictor}
\end{equation}
where $\phi^{PRE} \left(  \cdot  \right)$ denotes the function learned by the prediction layer and $h$ represents the predicted heatmaps.

\subsubsection{Auxiliary Decoder and Intermediate Supervision}
\label{subsubsec:auxdecoder}
Deep supervision \citep{lee2015deeply} refers to the technique that adds auxiliary supervision on some intermediate layers within a deep neural network. It facilitates multi-scale and multi-level feature learning by allowing error information back-propagation from multiple paths and alleviating the problem of vanishing gradients in deep neural networks. Leveraging the deep supervision idea, we also add an auxiliary decoder $\phi^{DEC}_{aux}\left(  \cdot  \right)$ after the penultimate stage of the encoder. Its structure is identical to $\phi^{DEC}\left(  \cdot  \right)$ described above. These two decoders do not share weights.

\subsection{Training objective}
\label{subsec:objective}
The ground truth heatmap is constructed by placing a Gaussian peak at each keypoint's location in the image plane. The number of heatmaps is identical to the number of keypoints predefined in the dataset, for example, 17 for the MS COCO dataset \citep{lin2014microsoft} and 14 for the AI Challenger dataset. We use an MSE loss to supervise the network during training. Mathematically, it is defined as:
\begin{equation}
L_{main} = \frac{1}{{H_K} {W_K} C}\sum\limits_{i,j,c} {{{\left\| {{h\left(i,j,c\right)} - h^{GT}\left(i,j,c\right)} \right\|}^2}},
\label{eq:L2_loss}
\end{equation}
where $C$ is the number of heatmaps, $i$, $j$, and $c$ are the spatial and channel index, $h$ and $h^{GT}$ are the predicted and ground truth heatmaps, respectively. Similar to Eq.~\eqref{eq:L2_loss}, an extra MSE loss $L_{aux}$ is also added to the auxiliary decoder as an intermediate supervision. The final training objective is defined as the weighted sum of both losses:
\begin{equation}
L = L_{main}  + \lambda {L_{aux}},
\label{eq:final_loss}
\end{equation}
where $\lambda$ is the weight for the auxiliary loss.

\section{Learning from diverse poses with efficient strategies}
\label{sec:learningStrategies}
CCM's capacity to model context information and learn discriminative feature representation can be exploited by learning from massive training samples with diverse poses. In this paper, we propose a hard-negative person detection mining strategy to migrate the mismatch problem of person detections during the training phase and testing phase, a joint-training strategy on unlabeled samples by knowledge distilling, and a joint-training strategy on external data with heterogeneous labels.

\subsection{Hard-Negative Person Detection Mining (HNDM)}
\label{subsec:hndm}
Top-down approaches detect person instances before detecting keypoints on them. On the one hand, although modern detection models have achieved a good detection performance, they may still produce some false positive detections due to occlusion, similar appearances, etc. On the other hand, the keypoint detection model is usually trained with ground truth bounding box annotations enclosing exact person instances. It has never seen any false positive detections during the training phase. Therefore, there is a mismatch between training and testing, which may lead to incorrect keypoint predictions for those false positive person detections. To address this issue, we propose a hard-negative person detection mining strategy.

\begin{figure}[t]
\begin{center}
   \includegraphics[width=1\linewidth]{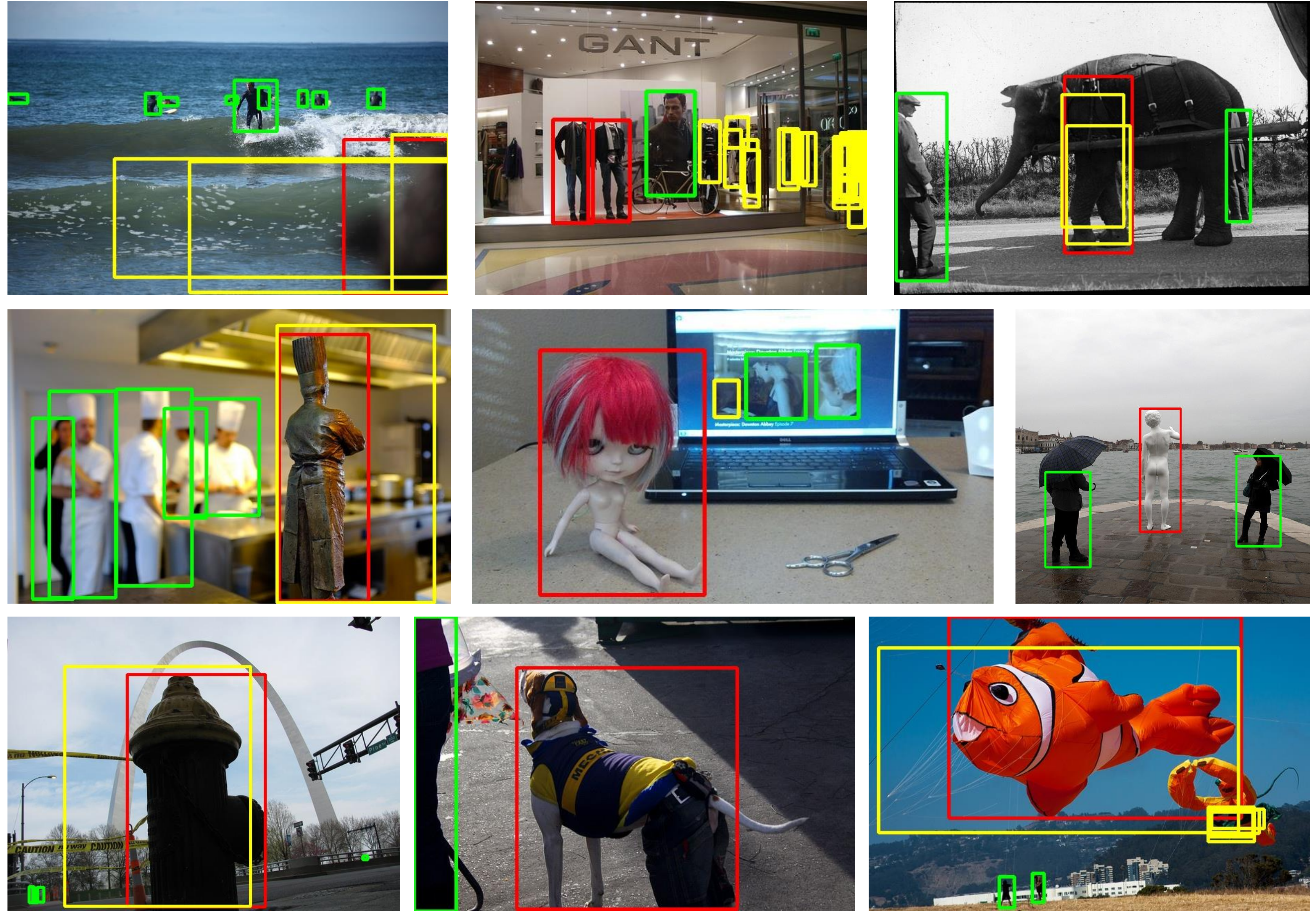}
\end{center}
   \caption{Illustration of the hard-negative detections. Green: ground truth person instances. Red: hard-negative detections. Yellow: low-score false positive detections.}
\label{fig:hndm}
\end{figure}

First, we trained a Mask R-CNN on the MS COCO training set containing only the category of person. ResNeXt152 was used as the backbone network. It achieved a mean average precision (AP) of 60.4 on the COCO minival dataset. Then, we evaluated the training set using the detection model and screened out those detections with sufficiently high scores, \emph{e.g.}, $ \ge0.5$, but no intersections with ground truth person instances. These were treated as hard-negative detections in this paper. Some examples are shown in Figure~\ref{fig:hndm}. During training CCM, these hard-negative detections could be added to the training set. Their keypoint heatmaps are set to all-zero maps. In this way, CCM learns to predict no keypoints on those false ``person instances''.

\subsection{Joint-Training on Unlabeled Samples by Knowledge Distilling}
\label{subsec:unlabel}
To increase pose diversity in the training samples, we leverage the MS COCO unlabeled dataset, which contains over 120k images. Since there are no person and keypoint annotations, we generate pseudo labels by referring to the knowledge distilling idea. First, we used the above-trained person detector to detect all possible person instances within the unlabeled images. Then, we screened out those detections with scores above a predefined threshold, which is determined by guaranteeing the number of average person instances per image to be identical to the one calculated from the ground truth annotations of the MS COCO training set. In our case, this threshold was 0.9924. Next, we trained several keypoint detection models using ResNet152 and HRNet-w48 as the backbone encoder network and used them to detect keypoints on the person detections obtained from the previous stage. We fused the predictions by different models, kept all keypoints with scores above 0.9 as the pseudo labels, and treated the rest as unlabeled. In this way, we distill the learned ``knowledge'' in the keypoint detection model to the pseudo labels of unlabeled training samples and use them to supervise the network.

\subsection{Joint-Training on External Data with Heterogeneous Labels}
\label{subsec:jointtraining}
Different datasets may not share the same annotation norms, even if they are used for the same purpose. For instance, 17 keypoints are used to define a human skeleton in the MS COCO dataset \citep{lin2014microsoft}, but only 14 in AI Challenger (AIC). Five keypoints correspond to the eyes, ears, and nose in MS COCO, while only the ``top of head'' is annotated in AIC. AIC has a keypoint annotation for the neck, which is absent in MS COCO. The other 12 keypoints corresponding to limbs are the same in both datasets. To use AIC with MS COCO, a common practice is to train a network on AIC and then change the number of channels in the final prediction layer and fine-tune this network on MS COCO. In this paper, we propose a simple but effective joint-training strategy that mixes the training samples in both datasets to leverage the diverse poses simultaneously. To make the training tractable, we align the labels in AIC with the ones in MS COCO by keeping the 12 common annotations and discarding the others.

To further adapt the trained model to the MS coco dataset, we can also add a finetune stage. In conclusion, the training strategies described above can be summarized as:
\begin{equation}
Train\left| \Phi  \right. \to Finetune\left| \Theta  \right.,
\label{eq:finetue}
\end{equation}
where $\Phi  \in \left\{ {A,AC,ACH,ACHU,CHU} \right\}$, $\Theta  \in \left\{ {C,CH} \right\}$, $A$ denotes the AIC training dataset, $C$ denotes the COCO training set, $H$ denotes the hard-negative training samples, and $U$ denotes the reprocessed unlabeled training set.

\section{Sub-pixel Refinement Techniques for Postprocessing}
\label{sec:subpixelrefinement}
To migrate the scale mismatch of representation between the high-resolution ground-truth keypoint location and the corresponding position on the low-resolution heatmap, different sub-pixel refinement techniques have been introduced, e.g., the 0.25-pixel shift of the maximum response pixel \citep{chen2018cascaded,xiao2018simple,sun2019deep,li2019rethinking}, the one-pixel shift of flipped heatmap \citep{xiao2018simple,sun2019deep}, and Gaussian filtering of predicted heatmap \citep{chen2018cascaded,li2019rethinking}. With this regard, we devise four sub-pixel refinement techniques by (1) exploring the second-order approximation to locate the maximum response at the sub-pixel level accuracy, (2) introducing the Soft Non-Maximum Suppression (Soft-NMS) to refine the maximum response's location instead of the original NMS, (3) extending the one-pixel shift of flipped heatmap to a general sub-pixel form, and (4) applying the Gaussian filter on predicted heatmaps. They are complementary (or orthogonal) to each other and can be conducted sequentially at different stages of the inference phase.

\subsection{Sub-pixel Refinement by the Second-Order Approximation}
\label{subsec:soa}
Sub-pixel refinement by the second-order approximation has ever been used in the depth super-resolution and disparity calculation literature \citep{yang2007spatial}. They use the cost volume at different disparities to find the optimal disparity that minimizes the inconsistency between the left and right views. Therefore, they face the issue to estimate the sub-pixel disparity given the costs at integer disparities. Likewise, we can adopt such a technique to estimate the sub-pixel keypoint location given the predicted heatmaps. In contrast to the one-dimensional cost volume, the heatmaps are in the two-dimensional space. Therefore, we present two kinds of second-order approximation by either using a parabola approximation for each dimension or using a paraboloid approximation simultaneously for the two dimensions.

\begin{figure}[t]
\centering
\includegraphics[width=0.6\linewidth]{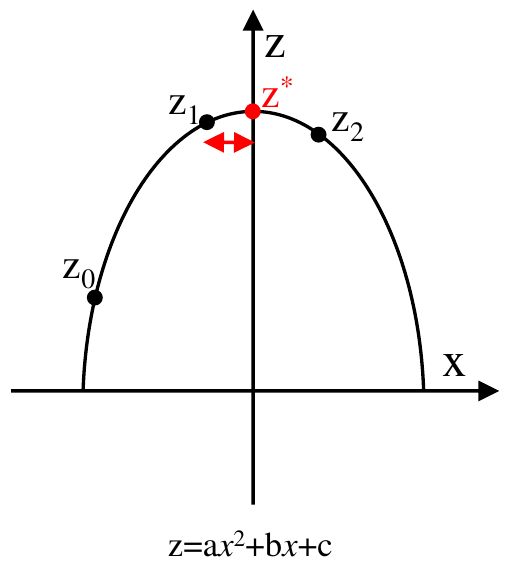}
\caption{The sub-pixel refinement by the second order approximation, i.e., the case of parabola.}
\label{fig:parabola}
\end{figure}

\subsubsection{The Parabola Approximation}
\label{subsec:parabola}
As we know that the heatmap is a Gaussian function w.r.t. the pixel dimension, nevertheless, it is reasonable to approximate it with a parabola function in a local neighborhood of the maximum pixel as shown in Figure~\ref{fig:parabola}:
\begin{equation}
z \left( x \right) = ax^2 + bx + c,
\label{eq:parabola}
\end{equation}
where $a$, $b$, and $c$ are the coefficients of the parabola. The maximum $z$ is subject to the first order condition:
\begin{equation}
\frac{{\partial z}}{{\partial x}} = 2ax + b = 0.
\label{eq:gradient}
\end{equation}
Therefore, the maximum $z$ is reached at $x^{*} =  - \frac{b}{{2a}}$.
Given $z_1$ is the maximum pixel at $x_0$, $z_0$ and the $z_2$ is the heatmap response at $x_{0}-1$ and $x_{0}+1$, we can calculate $x^{*}$ as:
\begin{equation}
{x^*} = x_0 + \frac{{{z_0} - {z_2}}}{{2\left( {{z_0} + {z_2} - 2{z_1}} \right)}}.
\label{eq:optimal}
\end{equation}
The second term in the right-hand side (RHS) of Eq.~\eqref{eq:optimal} is the sub-pixel shift from the detected maximum response pixel $x_0$ to the underlying maximum one $x^*$. Similarly, we can calculate the optimal $y^{*}$ along the vertical dimension.

\begin{figure}[t]
\centering
\includegraphics[width=0.58\linewidth]{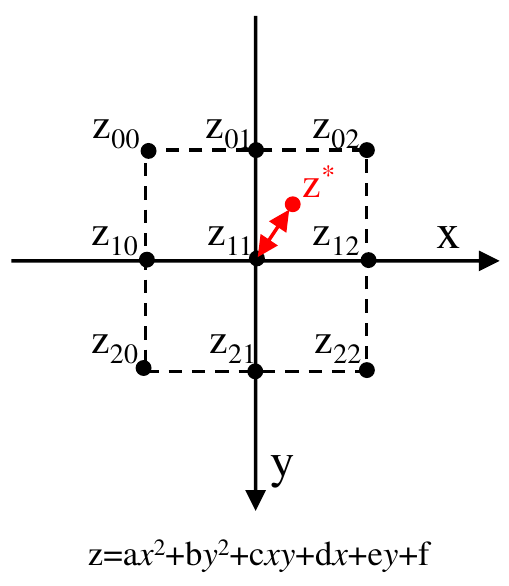}
\caption{The sub-pixel refinement by the second order approximation, i.e., the case of paraboloid.}
\label{fig:paraboloid}
\end{figure}

\subsubsection{The Paraboloid Approximation}
\label{subsec:paraboloid}
Since the heatmap is represented as a two-dimension Gaussian function, we can approximate it with a paraboloid function in a local neighborhood of the maximum pixel as shown in Figure~\ref{fig:paraboloid}:
\begin{equation}
z \left( x, y \right) = ax^2 + by^2 + cxy + dx + ey + f,
\label{eq:paraboloid}
\end{equation}
where $a$, $b$, $c$, $d$, $e$, and $f$ are the coefficients of the paraboloid. The maximum $z$ is subject to the first order condition:
\begin{equation}
\left\{ \begin{array}{l}
 \frac{{\partial z}}{{\partial x}} = 2ax + cy + d = 0 \\
 \frac{{\partial z}}{{\partial y}} = 2by + cx + e = 0 \\
 \end{array} \right..
\label{eq:gradient2}
\end{equation}
Therefore, the maximum $z$ is reached at:
\begin{equation}
\left\{ \begin{array}{l}
 {x^*} = \frac{{2bd - ce}}{{{c^2} - 4ab}} \\
 {y^*} = \frac{{2ae - cd}}{{{c^2} - 4ab}} \\
 \end{array} \right..
\label{eq:optimal2}
\end{equation}
As shown in Figure~\ref{fig:paraboloid}, $z_{11}$ is the maximum in the heatmap at $\left( x_0, y_0 \right)$, $z_{00} \sim z_{22}$ are its 8-neighbor heatmap responses. Assuming that the coordinate original is at $\left( x_0, y_0 \right)$, we can calculate the coefficients $a\sim e$ as:
\begin{equation}
\begin{array}{c}
 a = \frac{1}{8}\left[ {2\left( {{z_{12}} + {z_{10}} - 2{z_{11}}} \right)} \right. \\
  \qquad \quad + \left( {{z_{02}} + {z_{00}} - 2{z_{01}}} \right) \\
  \qquad \quad \ \ + \left. {\left( {{z_{22}} + {z_{20}} - 2{z_{21}}} \right)} \right] \\
 \end{array},
\label{eq:a}
\end{equation}
\begin{equation}
\begin{array}{c}
b = \frac{1}{8}\left[ {2\left( {{z_{01}} + {z_{21}} - 2{z_{11}}} \right)} \right. \\
  \qquad \quad + \left( {{z_{00}} + {z_{20}} - 2{z_{10}}} \right) \\
  \qquad \quad \ \ + \left. {\left( {{z_{02}} + {z_{22}} - 2{z_{12}}} \right)} \right] \\
 \end{array},
\label{eq:b}
\end{equation}
\begin{equation}
c = \frac{1}{4}\left( {{z_{00}} + {z_{22}} - {z_{02}} - {z_{20}}} \right),
\label{eq:c}
\end{equation}
\begin{equation}
d = \frac{1}{8}\left[ {\left( {{z_{02}} - {z_{00}}} \right) + \left( {{z_{22}} - {z_{20}}} \right) + 2\left( {{z_{12}} - {z_{10}}} \right)} \right],
\label{eq:d}
\end{equation}
\begin{equation}
e = \frac{1}{8}\left[ {\left( {{z_{20}} - {z_{00}}} \right) + \left( {{z_{22}} - {z_{02}}} \right) + 2\left( {{z_{21}} - {z_{01}}} \right)} \right],
\label{eq:e}
\end{equation}
Therefore, the maximum $z$ is reached at:
\begin{equation}
\left\{ \begin{array}{l}
 {x^*} = x_0 + \frac{{2bd - ce}}{{{c^2} - 4ab}} \\
 {y^*} = y_0 + \frac{{2ae - cd}}{{{c^2} - 4ab}} \\
 \end{array} \right..
\label{eq:optimal3}
\end{equation}
The second terms in the RHS of Eq.~\eqref{eq:optimal3} is the sub-pixel shift from the detected maximum response pixel $x_0$ (resp. $y_0$) to the underlying maximum one $x^*$ (resp. $y^*$). Given the heatmap, after locating the maximum pixel, we calculate the optimal location according to Eq.~\eqref{eq:optimal} or Eq.~\eqref{eq:optimal3}.

\subsection{Sub-pixel Refinement by Soft-NMS}
\label{subsec:softnms}

During the inference phase of top-down approaches, there may be several bounding boxes detected around a person instance. Consequently, we may estimate several poses on them. Similar to the Non-Maximum Suppression (NMS) postprocessing step used in object detection, NMS is also applied to the detected poses \citep{chen2018cascaded,xiao2018simple,sun2019deep}. Different from the Intersection over Union (IoU) used to compare the overlap between two bounding boxes, the Object Keypoint Similarity (OKS)-based IOU (OKS-IOU) is used to compare two poses. The original NMS is to filter out all the poses that have sufficient large OKS-IOUs with top-ranked detection. We argue that those poses can also be treated as reasonable estimates, which can be used to get a more stable result. Therefore, we present Soft-NMS for sub-pixel refinement as follows.

\begin{figure}[t]
\centering
\includegraphics[width=1\linewidth]{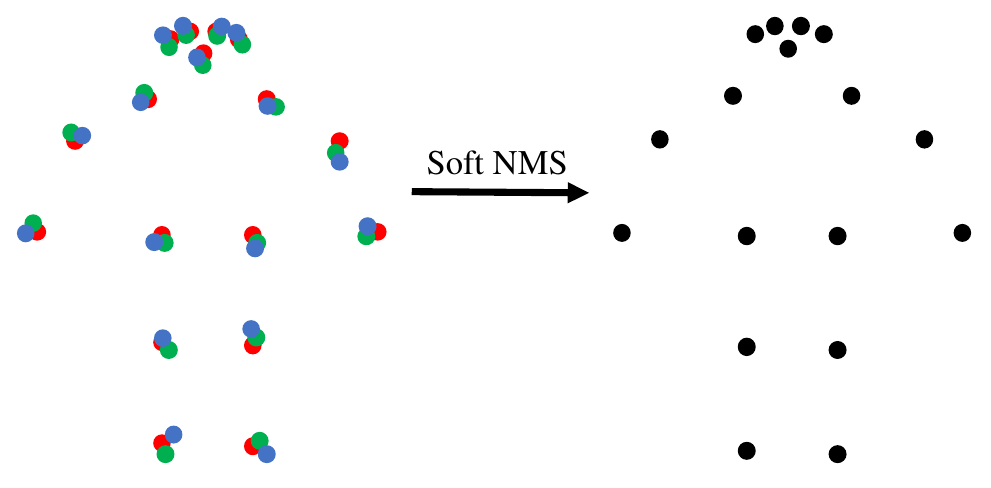}
\caption{The sub-pixel refinement by Soft NMS.}
\label{fig:softnms}
\end{figure}

As shown in Figure~\ref{fig:softnms}, the poses marked in red, blue, and green dots are three estimates and the red pose has the highest score. We can calculate the fusion results by leveraging the OKS-IOU as the fusion weight, i.e.,
\begin{equation}
p_i^* = \frac{{\sum\limits_{j \in {\Lambda _i}} {IOU_{ij}^{OKS}{p_j}} }}{{\sum\limits_{j \in {\Lambda _i}} {IOU_{ij}^{OKS}} }},
\label{eq:softnms}
\end{equation}
where ${{\Lambda _i}}$ is the index set of poses to be filtered given the top-ranked detection $p_i$, $IOU_{ij}^{OKS}$ is the OKS-IOU between $p_i$ and $p_j$. Note that we treat ${i \in {\Lambda _i}}$ and set $IOU_{ii}^{OKS} = 1$. We can re-write Eq.~\eqref{eq:softnms} as:
\begin{equation}
p_i^* = {p_i} + \frac{{\sum\limits_{j \in {\Lambda _i}} {IOU_{ij}^{OKS}\left( {{p_j} - {p_i}} \right)} }}{{\sum\limits_{j \in {\Lambda _i}} {IOU_{ij}^{OKS}} }},
\label{eq:softnms2}
\end{equation}
where the second term in the RHS is the sub-pixel shift from the top-ranked detection $p_i$ to the underlying best one $p_i^*$.

\subsection{Gaussian Filtering on Heatmaps}
\label{subsec:gb}
The regressed heatmap may not be as smooth as the ground truth Gaussian density map. A Gaussian filter-based postprocessing technique is proposed to smooth it and minimize the variance of the prediction \citep{chen2018cascaded,li2019rethinking}. We evaluate this technique and compare it with other sub-pixel refinement techniques.

\subsection{Sub-pixel Shift of Flipped Heatmaps}
\label{subsec:flipshit}

During the inference phase, some methods predict the pose from the flipped image and average the flipped heatmap with the original one to get the final prediction \citep{chen2018cascaded, xiao2018simple}. However, since the flipped heatmap is not aligned with the original one, one common practice is to shift the flipped heatmap by one pixel.

\begin{figure}[t]
\centering
\includegraphics[width=1\linewidth]{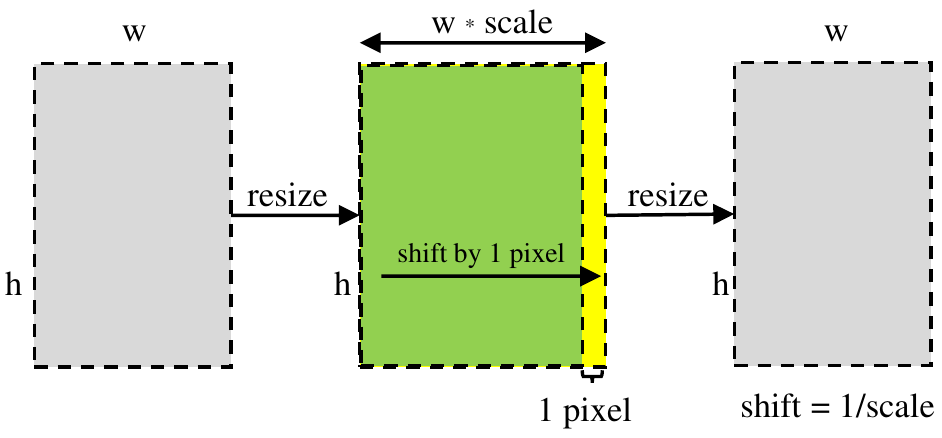}
\caption{The sub-pixel shift of flipped heatmaps.}
\label{fig:flipShift}
\end{figure}

In this paper, we extend the one-pixel shift to the sub-pixel shift as shown in Figure~\ref{fig:flipShift}. First, we resize the flipped heatmap by a scale ratio $r$ ($r\ge1$) along the horizontal axis. Then, we shift it by one pixel to the right. Then, we resize it back to the original size. As can be seen, the effective shift becomes $1/r$ pixel. We name it as the sub-pixel shift (SSP) in this paper. Note the above process is equivalent to a linear interpolation of the original heatmap and its one-pixel shifted version, i.e.,
\begin{equation}
h_{f}^*\left( {i,j} \right) = \left( {1 - \frac{1}{r}  } \right){h_{f}}\left( {i,j} \right) + \frac{1}{r} {h_{f}}\left( {i,j - j_0} \right),
\label{eq:shiftflip}
\end{equation}
where $h_{f}$ is the flipped heatmap, $j_0$ is the maximum shifted pixels, i.e., $j_0 = 1$ in this paper. $h_{f}^*$ is the sub-pixel estimate. It becomes the one-pixel shift technique when $r=1$.

\section{Experiments}
\label{sec:experiments}
We conducted extensive experiments to demonstrate the effectiveness of the proposed model. First, comprehensive ablation studies on the components of CM were presented, followed by the comparative studies of the proposed training strategies and sub-pixel refinement techniques. Next, we compared the proposed model with representative state-of-the-art methods in terms of detection accuracy, model complexity, and computational cost. Then, we presented some visual examples of the detection results by our model and explained the detection process by inspecting the learned features at each stage. Finally, we empirically studied the impact of visible and invisible annotations in our model and obtained useful some insights. 

\subsection{Experimental settings}
\label{subsec:settings}
\textbf{Datasets}: The COCO Keypoint Challenge addresses multi-person pose estimation in challenging uncontrolled conditions \citep{lin2014microsoft}. The dataset is split into training, minival, test-dev, and test-challenge sets. The training set includes 118k images and 150k person instances, the minival dataset includes 5000 images, and the test-dev set includes 20k images. It also provides an unlabeled dataset containing 123k images. 110k person instances and corresponding keypoints were detected using the method described in Section~\ref{subsec:unlabel}. The external dataset from AIC contains a training set with 237k images and 440k person instances and a validation set with 3000 images. We also evaluated CCM and the baseline model on the recently proposed OCHuman benchmark \citep{zhang2019pose} comprising heavily-occluded human instances to compare their performance on handling occluded cases. This dataset contains 8110 human instances with detailed keypoint annotations like COCO. It is divided into two subsets: OCHuman-Moderate and OCHuman-Hard. The first subset contains instances with MaxIoU in the range of 0.5 and 0.75, while the second contains instances with MaxIoU larger than 0.75. MaxIoU denotes the max IoU of a person with others in an image.

\textbf{Evaluation metrics}: We report the main results based on the object keypoint similarity (OKS)-based mean average precision (AP) over 10 OKS thresholds, where OKS defines the object keypoint similarity between different human poses. They are calculated as follows \citep{lin2014microsoft}:
\begin{equation}
AP = mean\left\{ {AP_{@\left( {0.50:0.05:0.95} \right)}} \right\},
\label{eq:mAP}
\end{equation}
\begin{equation}
AP_{@s} = \frac{{\sum\nolimits_p {\delta \left( {OK{S_p} > s} \right)} }}{{\sum\nolimits_p 1 }},
\label{eq:AP}
\end{equation}
\begin{equation}
OK{S_p} = \frac{{\sum\nolimits_i {\exp \left( {{{ - d_{pi}^2} \mathord{\left/
 {\vphantom {{ - d_{pi}^2} {\left( {2a_p^2\sigma _i^2} \right)}}} \right.
 \kern-\nulldelimiterspace} {\left( {2a_p^2\sigma _i^2} \right)}}} \right)\delta \left( {{v_{pi}} > 0} \right)} }}{{\sum\nolimits_i {\delta \left( {{v_{pi}} > 0} \right)} }},
\label{eq:oks}
\end{equation}
where $p$ is the person instance index, $i$ is the keypoint index, $\delta \left(  \cdot  \right)$ is the Kronecker function. $\delta \left(  \cdot  \right) = 1$ if the condition holds, otherwise 0. $s$ is a threshold, $d_{pi}$ is the Euclidean distance between the predicted $i^{th}$ keypoint of the person instance $p$ and its ground truth, $a_p$ is the area of the person instance $p$, $\sigma _i$ is the normalization factor predefined for each keypoint type, and $v_{pi}$ is the visible status.

\textbf{Implementation details}: The feature dimension $C_i$ of each CM was set to 256 for ResNet-50 and 128, 96, 64 for ResNet-152, 32 for HRNet-w32, and 48 for HRNet-w48. All backbone networks were pre-trained on the ImageNet dataset \citep{deng2009imagenet}. Gaussian initialization was used for convolutional and deconvolutional layers in the decoder. The weights and bias in BatchNorm layers were initialized as 1 and 0, respectively. CCM was implemented in Pytorch \citep{paszke2017automatic} and trained on four NVIDIA Tesla V100 GPUs using the Adam optimizer. Hyper-parameters were set by following \citep{xiao2018simple, sun2019deep}. We used the detection results on the minival set and test-dev set released in \citep{xiao2018simple} for a fair comparison, which were obtained by a faster-RCNN detector \citep{ren2015faster} with detection AP 56.4 for the person category on COCO val2017. We obtained the final predictions by averaging the heatmaps of the original and flipped image as in \citep{chen2018cascaded, newell2016stacked, xiao2018simple}.

\subsection{Ablation Studies}
\label{subsec:ablation}

\subsubsection{Ablation Study of the Components of CM}
\label{subsubsec:ablationmcm}

\begin{table}[htbp]
  \centering
  \caption{Ablation study on the components of CCM. AD: auxiliary decoder, D: dilated convolutions. AP/AR: mean average precision/recall on COCO minival set. Backbone network: ResNet-50 (R50) and HRNet-w32 (HR32).}
    %\begin{tabular}{ccccccc}
    \begin{tabular}{p{2.8cm}<{\centering}p{0.4cm}<{\centering}p{0.4cm}<{\centering}p{0.4cm}<{\centering}p{0.4cm}<{\centering}p{0.4cm}<{\centering}p{0.4cm}<{\centering}}
    \toprule
   Method & SE   & HDC    & AD  &D  & \emph{AP}    & \emph{AR} \\
    \midrule
   Baseline (R50)\\ \citep{xiao2018simple} &      &       &  &       & 70.4  & 76.3 \\
   \midrule
   \textbf{CCM (R50), K=3} &\checkmark  &  &   & & 71.7 & 77.9 \\
    &  & \checkmark      &       & & 71.8  & 78.0 \\
    &    &    &\checkmark       & & 72.1   & 78.2 \\
    &       &       &  & \checkmark     & 72.7  & 78.7 \\
    & \checkmark     & \checkmark     & \checkmark & \checkmark     & 73.5  & 79.1 \\
    \midrule
    \midrule
    HRNet-w32\\ \citep{sun2019deep} &  &  &   &  & 74.4    & 79.8 \\
   \textbf{CCM (HR32), K=1}  & \checkmark &\checkmark   &\checkmark  &  & {75.5}       & {80.9}  \\
   \textbf{CCM (HR32), K=2}  & \checkmark &\checkmark   &\checkmark  &  & {75.5}       & \textbf{81.0}  \\
   \textbf{CCM (HR32), K=3}  & \checkmark &\checkmark   &\checkmark  &  & \textbf{75.6}       & \textbf{81.0}  \\
    \bottomrule
    \end{tabular}%
  \label{tab:ablation_component}%
\end{table}%

First, we conducted an ablation study on the components of CM by training different variants on the COCO training set and calculating the AP and AR on the minival set. The backbone network was ResNet-50 (R50) and HRNet-w32, and the input size was $256 \times 192$. The results are shown in Table~\ref{tab:ablation_component}. For the ResNet-50 backbone, we used three CM modules in the decoder to keep consistent with the baseline model which used three deconvolutional layers. In this way, the feature map size increased from $8 \times 6$ at the end of the encoder to $64 \times 48$ at the end of the decoder. As can be seen, each component achieved gains over the baseline model \citep{xiao2018simple}, for example, adding the SE branch or HDC branch improved the detection accuracy by a margin of 1.3 AP or 1.4 AP over the baseline model. The auxiliary decoder and intermediate supervision also benefited the detection model and achieved a gain of 1.7 AP. Using dilated convolutions in the encoder could produce feature maps with $2\times$ higher resolution (i.e., $128 \times 96$), thereby benefiting the localization accuracy. As can be seen, it achieved a gain of 2.3 AP over the baseline model. These components are complementary to each other that the combination of them improved the detection accuracy further, i.e., a gain of 0.8$\sim$1.8 over the individual component.

For the HRNet-w32 backbone, since it could produce high-resolution features (i.e., $64 \times 48$), we only attached one CM module after HRNet-w32 and did not use any dilated convolutions, thereby increasing the feature map size to $128 \times 96$. Nonetheless, we also investigated whether extra CM modules were useful or not. To this end, we attached one or two extra CM modules (i.e., K=2 or 3) accordingly. To avoid huge computations for processing larger feature maps, we replaced the deconvolutional layers in the extra CM modules with convolutional layers to keep the feature map size. As can be seen from Table~\ref{tab:ablation_component}, CCM (K=1) outperformed the vanilla HRNet-w32 by a gain of 1.1 AP. Besides, adding extra CM modules only led to marginally better results while the parameters and computations increased from 25.56M and 7.92 GFLOPs (K=1) to 28.58M and 8.16 GFLOPs (K=2), and 28.6M and 8.4 GFLOPs (K=3), respectively. To make a trade-off between accuracy and computational efficiency, we chose K=1 as the default setting for the HRNet-w32 backbone.

\begin{table}[htbp]
%\footnotesize
  \centering
  \caption{Comparison of CCM and the Baseline model on the OCHuman dataset \citep{zhang2019pose}. Backbone network: ResNet-50 (R50).}
    %\begin{tabular}{ccccccc}
    \begin{tabular}{p{1.2cm}p{1.3cm}p{1cm}<{\centering}p{0.6cm}<{\centering}p{0.6cm}<{\centering}p{0.5cm}<{\centering}p{0.5cm}<{\centering}}
    \toprule
    Method    & Dataset & $AP$    & $AP^{@.5}$  & $AP^{@.75}$  & $AP^M$   & $AP^L$ \\
    \midrule
    \multirow{6}[0]{*}{Baseline} & $val$   & 59.9  & 79.3  & 66.1  & \textbf{64.7}  & 59.9 \\
          & $val_{\textbf{[0.5-0.75]}}$ & 62.7  & 81.5  & 69.4  & \textbf{64.7}  & 62.7 \\
          & $val_{\textbf{[0.75-1]}}$ & 42.9  & 63.1  & \textbf{46.4}  & -     & 42.9 \\
          \cmidrule{2-7}
          & $test$  & 52.1  & 70.6  & 57.0    & \textbf{72.6}  & 52.1 \\
          & $test_{\textbf{[0.5-0.75]}}$ & 61.1  & 80.4  & 66.7  & \textbf{89.2}  & 61.0 \\
          & $test_{\textbf{[0.75-1]}}$ & 43.4  & 60.8  & 47.4  & \textbf{40.0}    & 43.4 \\
    \midrule
    \midrule
    \multirow{6}[0]{*}{\textbf{CCM(R50)}} & $val$   & $\textbf{63.2}_{\textbf{(+3.3)}}$  & \textbf{81.9}  & \textbf{68.1}  & 53.7  & \textbf{63.2} \\
          & $val_{\textbf{[0.5-0.75]}}$ & $\textbf{66.2}_{\textbf{(+3.5)}}$  & \textbf{84.5}  & \textbf{71.8}  & 53.7  & \textbf{66.3} \\
          & $val_{\textbf{[0.75-1]}}$ & $\textbf{43.5}_{\textbf{(+0.6)}}$  & \textbf{66.0}    & 45.6  & -     & \textbf{43.5} \\
          \cmidrule{2-7}
          & $test$  & $\textbf{54.9}_{\textbf{(+2.8)}}$  & \textbf{74.5}  & \textbf{59.3}  & \textbf{72.6}  & \textbf{54.9} \\
          & $test_{\textbf{[0.5-0.75]}}$ & $\textbf{65.4}_{\textbf{(+4.3)}}$  & \textbf{83.7}  & \textbf{71.5}  & \textbf{89.2}  & \textbf{65.4} \\
          & $test_{\textbf{[0.75-1]}}$ & $\textbf{44.6}_{\textbf{(+1.2)}}$  & \textbf{65.0}    & \textbf{47.6}  & \textbf{40.0}    & \textbf{44.6} \\
    \bottomrule
    \end{tabular}%
  \label{tab:ochuman}%
\end{table}%

\begin{table}[htbp]
%\footnotesize
  \centering
  \caption{Comparison of CCM and the Baseline model on the PoseTrack validation set \citep{Andriluka2018PoseTrack}. Backbone network: ResNet-50 (R50) and HRNet-w32 (HR32).}
    %\begin{tabular}{ccccccc}
    \begin{tabular}{p{1.7cm}p{0.8cm}<{\centering}p{0.7cm}<{\centering}p{0.8cm}<{\centering}p{0.6cm}<{\centering}p{0.6cm}<{\centering}p{0.4cm}<{\centering}}
    \toprule
    Method    & $AP$    & $AP^{@.5}$  & $AP^{@.75}$  & $AP^M$   & $AP^L$ & $AR$ \\
    \midrule
    Baseline(R50)  & 69.6  & 86.4  & 75.7  & 40.6  & 73.8 & 72.2\\
    HRNet-w32 & 73.0  & 87.6  & 78.9  & 42.7  & 77.3 & 75.4\\
    \midrule
    \textbf{CCM(R50)} & $\textbf{71.3}_{\textbf{(+1.7)}}$  & \textbf{87.1}  & \textbf{76.5}  & \textbf{42.0}  & \textbf{75.5} & \textbf{74.1} \\
    \textbf{CCM(HR32)} & $\textbf{73.8}_{\textbf{(+0.8)}}$  & \textbf{87.7}  & \textbf{79.0}  & \textbf{44.8}  & \textbf{78.2} & \textbf{76.2} \\
    \bottomrule
    \end{tabular}%
  \label{tab:posetrack}%
\end{table}%

\begin{table*}[htbp]
%\small
  \centering
  \caption{Comparisons of CCM trained with the different strategies described in Section ~\ref{sec:learningStrategies}. A: the AIC training dataset; C: the COCO training set; H: the hard-negative training samples; U: the reprocessed unlabeled training set.}
    \begin{tabular}{lccccccccccc}
    %\begin{tabular}{p{3.3cm}p{0.35cm}<{\centering}p{0.5cm}<{\centering}p{0.6cm}<{\centering}p{0.35cm}<{\centering}p{0.35cm}<{\centering}p{0.35cm}<{\centering}}
    \toprule
      Method  & Training Strategy  & $AP$    & $AP^{@.5}$  & $AP^{@.75}$  & $AP^M$   & $AP^L$   & $AR$ & $AR^{@.5}$  & $AR^{@.75}$ & $AR^M$ & $AR^L$\\
    \midrule
    \multicolumn{1}{c}{\multirow{4}[0]{*}{\textbf{CCM (ResNet-50)}}} & A$\to$C &  74.8  & \textbf{90.6}  & 81.7  & 70.5  & 81.1  & 80.1  & \textbf{94.0}  & 86.1  & 75.7  & 86.5\\
    \multicolumn{1}{c}{} & AC$\to$C &  75.0  & 90.3  & 82.1  & 70.8  & 81.3  & 80.4  & \textbf{94.0}  & 86.7  & 76.0  & 86.6  \\
    \multicolumn{1}{c}{} & ACH$\to$CH &   75.3  & \textbf{90.6}  & 82.1  & 71.0  & 81.5  & 80.4  & \textbf{94.0}  & 86.5  & 76.0  & 86.6  \\
    \multicolumn{1}{c}{} & ACHU$\to$CH &  \textbf{75.6}  & 90.5  & \textbf{82.5}  & \textbf{71.4}  & \textbf{81.8}  & \textbf{80.7}  & \textbf{94.0}  & \textbf{86.8}  & \textbf{76.3}  & \textbf{87.0}  \\
    \bottomrule
    \end{tabular}%
  \label{tab:ablation_trainingStrategy}%
\end{table*}%

Besides, to compare the generalization ability of the proposed CCM and the baseline models when dealing with unseen heavily-occluded cases from other benchmarks, e.g., the OCHuman benchmark \citep{zhang2019pose} and the PoseTrack benchmark \citep{Andriluka2018PoseTrack}, we evaluated them in a zero-shot manner, where all models were trained on the COCO training set without further fine-tuning. It is noteworthy that we only include the results of the baseline models and our model for the following several reasons. Firstly, the goal is of this paper is to investigate the key factors that have strong impacts on the performance of human keypoint detection. Since it has already been evidenced by prior excellent work \citep{xiao2018simple,sun2019deep} that a better keypoint detector benefits pose tracking more effectively, we only focus on human keypoint detection in this paper. Here, we use the OCHuman dataset and PoseTrack dataset to validate the effectiveness of the proposed method for handling occluded cases. Secondly, the detection and tracking performance heavily depends on the person detector, but the detector as well as the implementation of joint propagation in \citep{xiao2018simple,sun2019deep} are not available. As a result, it will be unfair to compare with their results in a different setting. We plan to investigate the role of context for pose tracking in our future work, where spatial-, temporal- and channel-wise context could be exploited together to enhance feature representation. Thirdly, the Simple baseline method \citep{xiao2018simple} and the High-resolution Network method \citep{sun2019deep} are already two strong baseline models. It is noteworthy that many recent methods use HRNet as the backbone and have achieved SOTA performance on public datasets or challenges. Thereby, it is representative to compare our model with them on these two datasets. Besides, we use the zero-shot transfer setting to evaluate how the model trained on a representative dataset (e.g., MS COCO) generalizes to unseen samples from different data distributions, especially those samples containing large occlusions.

The results on the OCHuman benchmark are listed in Table \ref{tab:ochuman}. As can be seen, CCM outperformed the baseline model for handling occlusions by large margins, e.g., 3.3 points of AP gain on the validation set and 2.8 points of AP gain on the test set. The gains mainly arise from the occlusion cases $val_{[0.5-0.75]}$ and $test_{[0.5-0.75]}$, which contain occluded instances with MaxIoU in the range of 0.5 and 0.75. The results on the PoseTrack validation set are listed in Table \ref{tab:posetrack}. Our model outperforms the the baseline model with a ResNet-50 backbone and the HRNet-w32 model by 1.7 AP and 0.8 AP, confirming the superiority of CCM when dealing with heavily-occluded human instances. For a person instance with occluded body parts, it is challenging to detect both visible and invisible keypoints due to the self-occlusion, incomplete body, and perplexity with adjacent overlapped bodies. The proposed CM enables the detection model to learn discriminative feature representation for diverse poses and help it to recognize occluded keypoints. After the network sees diverse poses, it ``memorizes'' different poses with/without occlusions in the form of feature mapping. Inferring an occluded keypoint thus becomes easier by associating it with similar poses. More discussions will be presented in Section~\ref{subsec:subjective} and Section~\ref{subsec:insight}.

\textbf{Remarks}: 1) CM has better representative capacity than the plain deconvolution layer in the simple baseline method \citep{xiao2018simple} because it leverages spatial and channel context information explicitly; 2) CM is also complementary to the high-resolution module in \citep{sun2019deep} and improves the performance of the stronger HRNet-w32; and 3) CM effectively handles occlusions by learning context features to infer the occluded keypoints.

\begin{table}[htbp]
%\small
  \centering
  \caption{Study of the threshold setting in HNDM for the training strategy ``ACH$\to$CH'' described in Section ~\ref{subsec:hndm}.}
    \begin{tabular}{ccccccc}
    \toprule
    Threshold    & $AP$    & $AP^{@.5}$  & $AP^{@.75}$  & $AP^M$   & $AP^L$ & $AR$ \\
    \midrule
    0.5  & \textbf{75.3}  & 90.6  & 82.1  & 71.0  & \textbf{81.5}  & \textbf{80.4}\\
    0.7 & 75.2  & \textbf{90.9}  & \textbf{82.2}  & \textbf{71.1}  & 81.3 & \textbf{80.4} \\
    0.9 & 74.9 & 90.5  & 82.2  & 70.9  & 81.0 & 80.3 \\
    \bottomrule
    \end{tabular}%
  \label{tab:ablation_hnem_threshold}%
\end{table}%

\subsubsection{Comparison of Training Strategies}
\label{subsubsec:ablationstrategy}

\begin{table*}[htbp]
  \centering
  \small
  \caption{Experiments on different sub-pixel refinement techniques using the simple baseline method \citep{xiao2018simple}. SOA: sub-pixel refinement by the second-order approximation. Soft-NMS: Soft Non-Maximum Suppression. SSP: the sub-pixel shift of flipped heatmaps. GF: Gaussian filtering on predicted heatmaps. Backbone network: ResNet-50 (R50).}
    \begin{tabular}{cccccccccccccc}
    \toprule
      SOA &   Soft-NMS &   SSP &   GF     & $AP$    & $AP^{@.5}$  & $AP^{@.75}$ & $AP^M$ & $AP^L$ & $AR$  & $AR^{@.5}$  & $AR^{@.75}$ & $AR^M$ & $AR^L$ \\
    \midrule
    - & - & - & -  & 68.5  & 89.1  & 77.2  & 64.6  & 74.4  & 76.1  & 93.1  & 83.4  & 71.2  & 82.9  \\
    0.25 & - & - &  -  & 70.4  & 89.3  & 77.9  & 66.3  & 76.5  & 77.2  & 93.1  & 83.7  & 72.4  & 83.8  \\
    \gray
    parabola & - & - &  -  & 71.0  & 89.3  & 78.2  & 66.9  & 77.2  & 77.5  & 93.2  & 83.8  & 72.8  & 84.1  \\
    paraboloid & - & - &  -  & 71.0  & 89.3  & 78.5  & 66.8  & 77.2  & 77.5  & 93.1  & 84.0  & 72.8  & 84.1  \\
    \midrule
    \midrule
    - & \checkmark & - & -  & 69.3  & 89.1  & 77.3  & 65.2  & 75.4  & 76.5  & 93.1  & 83.4  & 71.6  & 83.3  \\
    0.25 & \checkmark & - &  -  & 70.9  & 89.3  & 78.0  & 66.7  & 77.2  & 77.4  & 93.1  & 83.7  & 72.6  & 84.1  \\
    \gray
    parabola & \checkmark & - &   - & 71.4  & 89.4  & 78.4  & 67.1  & 77.6  & 77.7  & 93.2  & 83.9  & 72.9  & 84.4  \\
    paraboloid & \checkmark & - & -   & 71.4  & 89.3  & 78.5  & 67.1  & 77.6  & 77.7  & 93.1  & 84.0  & 72.9  & 84.4  \\
    \midrule
    \midrule
    - & - & 1 & -  & 69.7  & 89.5  & 78.5  & 65.9  & 75.4  & 76.7  & 93.3  & 84.1  & 72.2  & 83.1  \\
    0.25 & - & 1 &  -  & 71.6  & 89.8  & 79.4  & 67.7  & 77.5  & 77.8  & 93.5  & 84.5  & 73.4  & 84.0  \\
    \gray
    parabola & - & 1 &   - & 72.1  & 89.8  & 79.7  & 68.2  & 78.1  & 78.1  & 93.5  & 84.7  & 73.8  & 84.3  \\
    paraboloid & - & 1 & -   & 72.2  & 89.7  & 79.7  & 68.2  & 78.2  & 78.2  & 93.4  & 84.7  & 73.8  & 84.4  \\
    \midrule
    \midrule
    - & - & - & \checkmark  & 68.8  & 88.9  & 77.5  & 64.9  & 74.7  & 76.2  & 93.0  & 83.6  & 71.3  & 82.9  \\
    0.25 & - & - & \checkmark  & 70.7  & 89.4  & 78.2  & 66.6  & 76.8  & 77.3  & 93.2  & 84.0  & 72.6  & 83.9  \\
    \gray
    parabola & - & - & \checkmark & 71.3  & 89.4  & 78.6  & 67.0  & 77.4  & 77.6  & 93.1  & 84.0  & 72.9  & 84.1  \\
    paraboloid & - & - & \checkmark  & 71.2  & 89.4  & 78.5  & 67.0  & 77.4  & 77.6  & 93.2  & 84.0  & 72.9  & 84.1  \\
    \midrule
    \midrule
    - & \checkmark & 1 & \checkmark  & 70.8  & 89.7  & 78.8  & 66.9  & 76.6  & 77.2  & 93.5  & 84.2  & 72.8  & 83.5  \\
    0.25 & \checkmark & 1 & \checkmark  & 72.2  & 89.9  & 79.6  & 68.3  & 78.2  & 78.1  & 93.6  & 84.6  & 73.7  & 84.2  \\
    \gray
    parabola & \checkmark & 1 & \checkmark & 72.6  & 89.9  & \textbf{79.8}  & 68.6  & 78.5  & 78.3  & 93.5  & 84.7  & 74.0  & 84.4  \\
    paraboloid & \checkmark & 1 & \checkmark  & 72.6  & 89.9  & 79.7  & 68.6  & 78.5  & 78.3  & 93.5  & 84.6  & 74.0  & 84.4  \\
    \midrule
    \midrule
    \gray
    parabola & \checkmark & 0.8 & \checkmark  & \textbf{72.8}  & \textbf{89.9}  & 79.7  & \textbf{68.7}  & 78.9  & \textbf{78.5}  & \textbf{93.6}  & \textbf{84.7}  & \textbf{74.1}  & 84.7  \\
    parabola & \checkmark & 0.6 & \checkmark  & 72.7  & 89.9  & 79.7  & 68.4  & \textbf{79.1}  & 78.4  & 93.5  & 84.7  & 73.8  & \textbf{84.9}  \\
    parabola & \checkmark & 0.4 & \checkmark  & 72.3  & 89.9  & 79.5  & 67.9  & 78.9  & 78.1  & 93.5  & 84.6  & 73.4  & 84.8  \\
    parabola & \checkmark & 0.2 & \checkmark  & 71.7  & 89.9  & 79.2  & 67.1  & 78.3  & 77.5  & 93.4  & 84.3  & 72.7  & 84.3  \\
    parabola & \checkmark & 0 & \checkmark  & 70.7  & 89.8  & 78.6  & 66.1  & 77.5  & 76.7  & 93.4  & 84.0  & 71.7  & 83.7  \\

    \bottomrule
    \end{tabular}%
  \label{tab:resnet50}%
\end{table*}%

Next, we present the results of using different training strategies described in Section ~\ref{sec:learningStrategies} in Table~\ref{tab:ablation_trainingStrategy}, where A$\to$C denotes the transfer learning strategy, AC$\to$C denotes the joint-training strategy, i.e., training CCM on both AIC and COCO datasets then fine-tuning it on the COCO dataset. Other symbols have a similar meaning. As can be seen, leveraging the external AIC dataset increased the AP by 1.3 compared with the model trained on the COCO dataset in Table~\ref{tab:ablation_component}. The improvement became 1.5 AP when using the proposed joint-training strategy. The proposed HNDM method increased the AP further by an extra gain of 0.3, while the AR remained the same. This is reasonable since HNDM aims to suppress the keypoints of the false-positive person detections, meaning that it can increase the precision but has little influence on the recall. After exploiting the unlabeled dataset, CCM obtained a final AP of 75.6, a gain of 2.1 over the same model trained on the COCO dataset. 

We also evaluated the impact of the detection score threshold in HNDM. Specifically, we set its value to 0.5, 0.7, and 0.9 in the training strategy ``ACH$\to$CH''. The results are summarized in Table~\ref{tab:ablation_hnem_threshold}. As can be seen, the performance decreased with the increase of the threshold. Note that fewer hard negative detections were included in the training set when a larger threshold was used. For example, there were 11,762 detections at the threshold of 0.5 while only 4,359 and 952 detections were left at the threshold of 0.7 and 0.9, respectively. On the one hand, since there were many person detection proposals with low scores from the person detectors to increase the recall, thereby leveraging hard negative person detections with low scores (e.g., $0.5\sim0.7$) and predicting all-zero heatmaps could reduce false positive keypoints on those person detection proposals. On the other hand, due to the annotation policy, human-like objects such as model, sculpture, and doll (please see Figure~\ref{fig:hndm}) were not annotated as the person category, which however could be detected with high scores by the person detector. Since they shared similar appearance with real human bodies, only using these hard negative person detections with high scores (e.g., $\geq0.9$) increased the difficulty for learning discriminative feature representation for keypoint detection.

\textbf{Remarks}: 1) The proposed joint-training strategy is more effective than transfer learning by reducing the domain gap during the pretraining phase; 2) HNDM can deal with the false-positive person detections, thereby improving the detection precision; and 3) exploiting extra unlabeled data using knowledge distilling enables the network to learn more discriminative features from abundant and diverse samples. 

\subsubsection{Comparison of Sub-pixel Refinement Techniques}
\label{subsubsec:ablationsubpixel}
We conducted the contrastive experiments by using different sub-pixel refinement techniques in both the simple baseline method \citep{xiao2018simple} and the High-Resolution Network. The results are summarized in Table~\ref{tab:resnet50} and Table~\ref{tab:hrnet32}.

\begin{table*}[htbp]
  \centering
  \small
  \caption{Experiments on different settings of sub-pixel refinement tricks using the High-Resolution Network \citep{sun2019deep} (HRNet-w32). SOA: sub-pixel refinement by the second-order approximation. Soft-NMS: sub-pixel refinement by Soft Non-Maximum Suppression (Soft-NMS). SSP: the sub-pixel shift of flipped heatmaps. GF: Gaussian filtering on heatmaps.}
    \begin{tabular}{cccccccccccccc}
    \toprule
      SOA &   Soft-NMS &   SSP &   GF     & $AP$    & $AP^{@.5}$  & $AP^{@.75}$ & $AP^M$ & $AP^L$ & $AR$  & $AR^{@.5}$  & $AR^{@.75}$ & $AR^M$ & $AR^L$ \\
    \midrule
    - & - & - & -  & 71.7  & 90.0    & 80.0    & 67.9  & 77.6  & 78.8  & 94.0    & 85.4  & 74.2  & 85.1 \\
    0.25 & - & - &  -  & 73.8  & 90.4  & 80.6  & 69.7  & 79.9  & 80.0    & 94.0    & 85.9  & 75.6  & 86.3 \\
    \gray
    parabola & - & - &  -  & 74.3  & 90.4  & 81.1  & 70.3  & 80.4  & 80.4  & 94.0    & 86.2  & 76.0    & 86.5 \\
    paraboloid & - & - &  -  & 74.3  & 90.4  & 81.2  & 70.3  & 80.4  & 80.4  & 94.1  & 86.2  & 76.0    & 86.5 \\
    \midrule
    \midrule
    - & \checkmark & - & -  & 72.6  & 90.0    & 80.0    & 68.5  & 78.7  & 79.2  & 94.0    & 85.5  & 74.6  & 85.6 \\
    0.25 & \checkmark & - &  -  & 74.3  & 90.4  & 80.7  & 70.1  & 80.5  & 80.3  & 94.0    & 86.0    & 75.8  & 86.6 \\
    \gray
    parabola & \checkmark & - &   - & 74.7  & 90.4  & 81.2  & 70.5  & 80.9  & 80.5  & 94.0    & 86.2  & 76.1  & 86.7 \\
    paraboloid & \checkmark & - & -   & 74.7  & 90.4  & 81.3  & 70.5  & 80.9  & 80.6  & 94.1  & 86.3  & 76.1  & 86.8 \\
    \midrule
    \midrule
    - & - & 1 & -  & 72.7  & 90.6  & 81.1  & 68.8  & 78.6  & 79.4  & 94.3  & 86.3  & 75.0    & 85.6 \\
    0.25 & - & 1 &  -  & 74.7  & 90.7  & 82.3  & 70.6  & 80.8  & 80.5  & 94.3  & 87.1  & 76.1  & 86.6 \\
    \gray
    parabola & - & 1 &   - & 75.2  & 90.8  & 82.5  & 71.1  & 81.3  & 80.8  & 94.3  & 87.1  & 76.5  & 86.8 \\
    paraboloid & - & 1 & -   & 75.2  & 90.8  & 82.4  & 71.1  & 81.3  & 80.8  & 94.3  & 87.0    & 76.5  & 86.8 \\
    \midrule
    \midrule
    - & - & - & \checkmark  & 72.3  & 90.3  & 80.3  & 68.4  & 78.3  & 79.1  & 94.1  & 85.6  & 74.5  & 85.5 \\
    0.25 & - & - & \checkmark  & 74.3  & 90.4  & 81.2  & 70.1  & 80.5  & 80.2  & 94.0    & 86.1  & 75.8  & 86.4 \\
    \gray
    parabola & - & - & \checkmark & 74.8  & 90.4  & 81.4  & 70.7  & 81.0    & 80.5  & 94.0    & 86.1  & 76.2  & 86.7 \\
    paraboloid & - & - & \checkmark  & 74.7  & 90.5  & 81.4  & 70.7  & 80.9  & 80.5  & 94.0    & 86.2  & 76.2  & 86.6 \\
    \midrule
    \midrule
    - & \checkmark & 1 & \checkmark  & 74.2  & 90.6  & 81.5  & 70.1  & 80.3  & 80.1  & 94.2  & 86.5  & 75.7  & 86.3 \\
    0.25 & \checkmark & 1 & \checkmark  & 75.5  & 90.8  & 82.6  & 71.4  & 81.7  & 80.9  & 94.3  & 87.2  & 76.7  & 86.9 \\
    \gray
    parabola & \checkmark & 1 & \checkmark & 75.7  & 90.8  & 82.8  & 71.6  & 81.9  & 81.1  & 94.3  & 87.3  & 76.8  & 87.1 \\
    paraboloid & \checkmark & 1 & \checkmark  & 75.7  & 90.8  & 82.7  & 71.6  & 81.9  & 81.1  & 94.3  & 87.3  & 76.8  & 87.1 \\
   \midrule
    \midrule
    \gray
    parabola & \checkmark & 0.8 & \checkmark  & \textbf{76.0}  & 90.8  & 82.8  & \textbf{71.8}  & 82.3  & \textbf{81.3}  & \textbf{94.3}  & \textbf{87.3}  & \textbf{77.0}  & 87.4  \\
    parabola & \checkmark & 0.6 & \checkmark  & 76.0  & \textbf{90.9}  & \textbf{82.9}  & 71.6  & \textbf{82.5}  & 81.3  & 94.3  & 87.3  & 76.9  & \textbf{87.5}  \\
    parabola & \checkmark & 0.4 & \checkmark  & 75.6  & 90.8  & 82.7  & 71.1  & 82.2  & 80.9  & 94.2  & 87.2  & 76.4  & 87.3  \\
    parabola & \checkmark & 0.2 & \checkmark  & 74.9  & 90.8  & 82.2  & 70.3  & 81.6  & 80.3  & 94.2  & 87.0  & 75.7  & 86.9  \\
    parabola & \checkmark & 0 & \checkmark  & 73.8  & 90.7  & 81.8  & 69.2  & 80.7  & 79.4  & 94.2  & 86.6  & 74.6  & 86.2  \\
    \bottomrule
    \end{tabular}%
  \label{tab:hrnet32}%
\end{table*}%

First, the sub-pixel refinement techniques by second-order approximation described in Section~\ref{subsec:soa} consistently improved the performance of both methods. Note that shifting towards the gradient directions by 0.25 pixels was effective and improved the performance of ResNet-50 and HRNet-w32 by 1.9 AP and 2.1 AP, respectively. Nevertheless, it only used the first-order derivative information and the shift was a fixed value, limiting its performance. In contrast, the proposed SOA refinement further increased the AP from 68.5 to 71.0, and 71.7 to 74.3, for ResNet-50 and HRNet-w32, respectively. With the second-order approximation, SOA could adaptively calculate the shift vector for each heatmap. The paraboloid-based SOA and parabola-based SOA performed similarly. It is reasonable because the target heatmap is a 2D Gaussian density map where the density along the x-axis is independent of the density along the y-axis. The predicted heatmap had a similar pattern. Therefore, the paraboloid-based SOA had no obvious advantage over the parabola-based SOA. Nevertheless, the paraboloid-based SOA may be useful for those scenarios where the joint densities along different axes are not independent of each other, i.e., there is an elliptical response with a bias direction in the heatmap.

Second, Soft-NMS was effective, which improved the performance by 0.8 AP and 0.9 AP for ResNet-50 and HRNet-w32, respectively. The weighted fusion defined by Eq.~\eqref{eq:softnms} shifted the keypoint locations in sub-pixels by considering the reasonable estimations rather than filtering them out as done in standard NMS. Besides, the SOA technique was complementary to Soft-NMS. For example, the parabola-based SOA technique combined with Soft-NMS improved the performance further by 0.4 AP compared with using the parabola-based SOA technique individually and improved the performance further by 2.1 AP compared with using Soft-NMS individually for both baselines. 

Third, the flip test together with the one-pixel shift of flipped heatmaps improved the performance consistently, i.e., by a margin of 1.2 AP and 1.0 AP ResNet-50 and HRNet-w32, respectively. The SOA technique was complementary to the flip test. We leave the analysis on the sub-pixel shift later.

Fourth, Gaussian filtering was beneficial for improving detection performance. A gain of 0.3 AP and 0.6 AP was achieved for ResNet-50 and HRNet-w32, respectively. It was complementary to the SOA technique. Using them together outperformed using each of them individually. To show the complementarity among all the techniques, we used them together in both methods. They boosted the vanilla baseline's performance by a large margin, e.g., 4.1 AP for ResNet-50 and 4.0 AP for HRNet-w32.

We evaluated the influence of the shifted pixel for the flipped heatmap described in Section~\ref{subsec:flipshit}. As can be seen from the bottom rows in Table~\ref{tab:resnet50} and Table~\ref{tab:hrnet32}, the performance dropped significantly without shifting the flipped heatmap, i.e., from 72.6 AP to 70.7 AP for ResNet-50, and from 75.7 AP to 73.8 AP for HRNet-w32, since the flipped heatmap was not aligned with the original one. Generally, increasing the shifted pixel from zero to 0.8 consistently improved the detection accuracy. It saturated at a shift of 0.8 pixels and then dropped at a shift of one pixel. We used the sub-pixel refinement techniques in our submission to the 2019 COCO Keypoint Detection Challenge.

\textbf{Remarks}: 1) Each sub-pixel refinement technique has a positive but slightly different influence on the performance, i.e., SOA $\ge$ SSP $\ge$ Soft-NMS $\ge$ GF; 2) the proposed SOA and SSP are much better than their vanilla counterparts due to the closed-form and sub-pixel level approximation; and 3) these techniques are complementary since they are carried out in subsequent steps for unique and explicit purposes.

\begin{table}[htbp]
  \centering
  \caption{Inference time (millisecond, i.e., ms) of different sub-pixel refinement techniques and the proposed CM module. Network Forward: the forward process of different networks. Map $\rightarrow$ Coord: the process of getting final keypoint coordinates from predicted heatmaps. Evaluation: the process of calculating the evaluation metrics of the keypoint detection results. Backbone network: HRNet-w32.}
    %\begin{tabular}{cccccccc}
    \begin{tabular}{p{0.8cm}<{\centering}p{0.8cm}<{\centering}p{0.3cm}<{\centering}p{0.3cm}<{\centering}p{0.2cm}<{\centering}p{1.1cm}<{\centering}p{1cm}<{\centering}p{0.5cm}<{\centering}}
    \toprule
   Model & SOA   & Soft-NMS & SSP  & GF  & \emph{Network Forward}    & \emph{Map $\rightarrow$ Coord} & Eval-uation \\
   \midrule
   \multirow{5}[0]{*}{Baseline} & parabola  & \checkmark & 0.4 & \checkmark & 2.44 & 4.99 & 0.43\\
    & parabola  & \checkmark & 1 & \checkmark & \textbf{2.43}  & 4.73 & 0.43 \\
    & parabola  & \checkmark & 1 & - & 2.45   & 4.13 & 0.43 \\
    & 0.25  & \checkmark & 1 & - & 2.46  & 3.85 & 0.43\\
    & 0.25  & - & 1 & - & 2.45  & \textbf{3.81} & 0.41 \\
    \midrule
   \multirow{2}[0]{*}{\textbf{CCM}} &  parabola  & \checkmark & 0.4 & \checkmark & 2.50  & 9.30 & \textbf{0.39} \\
    & 0.25  & - & 1 & - & 2.51  & 7.53 & \textbf{0.39}\\
    \bottomrule
    \end{tabular}%
  \label{tab:inference_time}%
\end{table}%

\begin{table*}[htbp]
%\small
  \centering
  \caption{Comparisons of CAPE-Net and SOTA methods on the COCO minival set.}
    \begin{tabular}{llccccccccc}
    %\begin{tabular}{p{3cm}p{0.4cm}<{\centering}p{0.6cm}<{\centering}p{0.6cm}<{\centering}p{0.4cm}<{\centering}p{0.4cm}<{\centering}p{0.4cm}<{\centering}}
    \toprule
      Method    & \multicolumn{2}{c}{Backbone} & \#Params & GFLOPs &$AP$    & $AP^{@.5}$  & $AP^{@.75}$  & $AP^M$   & $AP^L$   & $AR$ \\
    \midrule
     & &  & & \multicolumn{7}{c}{Input size: $256\times192$} \\
    Baseline \citep{xiao2018simple} & \multicolumn{1}{c}{\multirow{4}[0]{*}{Small}} & ShuffleNet-v2 & 8.78M & 4.07 & 63.9    &  86.6     &  71.4     &  60.8     &  70.1     & 70.3 \\
    Baseline \citep{xiao2018simple} & \multicolumn{1}{c}{} & MobileNet-v2 & 9.57M & 4.17 & 64.3     &  86.3     &  72.2     &  60.9     &  70.9     & 70.5 \\
    \textbf{CCM} & \multicolumn{1}{c}{} & ShuffleNet-v2 & 9.72M & 4.75 & {67.9}     &  \textbf{88.4}     &  {75.3}     &  {63.8}     &  {74.0}    & {74.2}\\
    \textbf{CCM} & \multicolumn{1}{c}{} & MobileNet-v2 & 10.1M & 4.80 & \textbf{68.4}     &  {88.0}     &  \textbf{75.7}     &  \textbf{64.2}     &  \textbf{74.5}    & \textbf{74.3} \\
    \midrule
    \midrule
    & &  & & \multicolumn{7}{c}{Input size: $256\times192$} \\
    Hourglass \citep{newell2016stacked} & \multicolumn{1}{c}{\multirow{8}[0]{*}{Medium}} & 8xHourglass & 25.1M & 14.3 &  66.9     &   -    &    -   &    -   &    -   & - \\
    CPN \citep{chen2018cascaded} & \multicolumn{1}{c}{}  & ResNet-50 & 27.0M & 6.20 & 69.4     &  -     &   -    &   -    &   -    & - \\
    Baseline \citep{xiao2018simple} & \multicolumn{1}{c}{}  & ResNet-50 & 34.0M & 8.20 & 70.4     &  88.6     &  78.3     &  67.1     &  77.2     & 76.3 \\
    Baseline \citep{xiao2018simple} & \multicolumn{1}{c}{}  & ResNeXt-50 & 33.5M & 8.35 & 70.6     &  88.9     &  77.9    &  67.2     &  77.5     & 76.5 \\
    HRNet \citep{sun2019deep} & \multicolumn{1}{c}{}  & HRNet-w32 & 28.5M & 7.68 & 74.4     &  90.5     &  81.9     &  70.8     &  81.0    & 79.8 \\
    \textbf{CCM} & \multicolumn{1}{c}{}  & ResNet-50 & 40.7M & 45.4 & 74.3         & 90.3         & 81.3         &  70.0        &  80.6       & 79.6  \\
    \textbf{CCM} & \multicolumn{1}{c}{}  & ResNeXt-50 & 40.1M & 45.4 & 74.5         & 90.4         & 81.2         &  70.0        &  81.0       & 79.6  \\
    \textbf{CCM} & \multicolumn{1}{c}{}  & HRNet-w32 & 28.6M & 7.92 & \textbf{76.7}         & \textbf{91.1}         & \textbf{83.5}         &  \textbf{72.5}        &  \textbf{83.0}       & \textbf{81.8}  \\
    \midrule
    \midrule
    & &  & & \multicolumn{7}{c}{Input size: $384\times288$} \\
    Baseline \citep{xiao2018simple} & \multicolumn{1}{c}{\multirow{4}[0]{*}{Large}} & ResNet-152 & 68.6M & 35.9 & 74.3     &  89.6     &  81.1     &  70.5     &  79.7     & 79.7 \\
    HRNet \citep{sun2019deep} & \multicolumn{1}{c}{} & HRNet-w48 & 63.6M & 35.4 & 76.3     &  90.8     &  82.9     &  {72.3}     &  {83.4}    & 81.2 \\
    \textbf{CCM} & \multicolumn{1}{c}{} & ResNet-152 & 63.5M & 40.1 & {76.7}     &  {91.2}     &  {83.4}     &  {72.4}     &  83.2    & {81.7} \\
    \textbf{CCM} & \multicolumn{1}{c}{} & HRNet-w48 & 63.7M & 36.6 & \textbf{77.5}     &  \textbf{91.2}     &  \textbf{83.6}     &  \textbf{73.0}     &  \textbf{84.0}    & \textbf{82.3} \\
    \bottomrule
    \end{tabular}%
  \label{tab:SOTA_minival}%
\end{table*}%

\subsubsection{Inference Time Analysis}
\label{subsubsec:inferencetime}
We also compared the inference time of different sub-pixel refinement techniques and the proposed CM module. Specifically, we chose HRNet-w32 as the backbone network. We divided the inference process into three parts, i.e., 1) the network forward process (denoting ``Network Forward''); 2) the process of getting final keypoint coordinates from predicted heatmaps (denoting ``Map $\rightarrow$ Coord''); and 3) the process of calculating the evaluation metrics of the keypoint detection results (denoting ``Evaluation''), and recorded their inference time separately. For each setting, we ran the model three times and calculated the average inference time for single person instance. The results are summarized in Table~\ref{tab:inference_time}.

As can be seen, the inference time of different post-processing techniques mainly differs in the process of ``Map $\rightarrow$ Coord''. For example, replacing the NMS in the default setting of the baseline model \citep{xiao2018simple} with the proposed soft-NMS only increased by 0.04 millisecond (ms). SOA and SSP increased the inference time slightly, i.e., from 3.85 ms to 4.13 ms and from 4.73 ms to 4.99 ms. However, Gaussian filtering required much more computations and increased the inference time from 4.13 ms to 4.73 ms. It is noteworthy that the process of ``Map $\rightarrow$ Coord'' cost more inference time than ``Network Forward'' since it was mainly carried out on CPU while network forward computation was carried on GPU. Adding a CM module on HRNet-w32 only increased the network forward time slightly, i.e., about 0.05 ms. However, since the heatmaps generated by CCM were two times larger than those by the baseline model, the process of ``Map $\rightarrow$ Coord'' became much slower, i.e., about two times. It is worth further study to find a fast GPU implementation for this process.

\subsection{Comparison with state-of-the-art methods}
\label{subsec:mainResults}
The results of CCM and SOTA methods on the COCO minival are summarized in Table~\ref{tab:SOTA_minival}. CCM was trained on the COCO dataset without using the external AI Challenger dataset. We evaluated the proposed approach on three groups of backbone networks, i.e., the small ones including ShuffleNet-v2 and MobileNet-v2, the medium ones including ResNet-50 and HRNet-w32, and the large ones including ResNet-152 and HRNet-w48, respectively. As can be seen, our small model based on ShuffleNet-v2 and MobileNet-v2 significantly improved the detection accuracy compared with the baseline model \citep{xiao2018simple}, i.e., a gain of 4.0 AP and 4.1 AP, respectively. The improvement is at the cost of 5\% $\sim$ 10\% more parameters and about 15\% more GFLOPs, which are affordable. CCM also outperformed the Hourglass model \citep{newell2016stacked} and was comparable with the CPN \citep{chen2018cascaded} based on ResNet-50. 

As for the medium backbone networks, simple baseline method \citep{xiao2018simple} achieved similar performance using ResNet-50 and ResNeXt-50, but they were inferior to the High-Resolution Network \citep{sun2019deep} based on HRNet-w32. The proposed CCM based on ResNet-50 achieved a 74.3 AP and outperformed other models with the same input size and backbone network, for example, a gain of 3.9 AP over the simple baseline method \citep{xiao2018simple}. Replacing ResNet-50 to ResNeXt-50 leads to a slightly better result, i.e., from 74.3 AP to 74.5 AP. When using the HRNet-w32 as the backbone network, our CCM model increased the AP from 74.4 to 76.7 and achieved the best performance among all the models. Note that CCM based on ResNet-50 or ResNeXt-50 has more parameters and GFLOPs than the baseline model since it uses dilated convolutions to increase the feature map size and three CMs in the decoder. As for the one based on HRNet-w32, it has roughly the same amount of parameters and GFLOPs as its counterpart since only one CM was used.

\begin{table*}[htbp]
%\small
  \centering
  \caption{Comparisons of CCM and SOTA methods on the COCO test-dev set. Input size: $353\times257$ for G-RMI; $320\times256$ for RMPE; $384\times288$ for CPN, Baseline, HRNet, MSPN, DARK, RSN, and CCM. The symbol ``*'' denotes external data, ``+'' denotes ensemble models, ``$\dagger$'' and ``$\ddagger$'' denote the champion of the 2018 and 2019 COCO Keypoint Challenge, respectively.}
    \begin{tabular}{llcccccccc}
    %\begin{tabular}{p{3.3cm}p{0.35cm}<{\centering}p{0.5cm}<{\centering}p{0.6cm}<{\centering}p{0.35cm}<{\centering}p{0.35cm}<{\centering}p{0.35cm}<{\centering}}
    \toprule
      Method  & Backbone & \#Params & GFLOPs & $AP$    & $AP^{@.5}$  & $AP^{@.75}$  & $AP^M$   & $AP^L$   & $AR$ \\
    \midrule
    Mask-RCNN \citep{he2017mask} & ResNet-50 &  - & - & 63.1    &  87.3     &  68.7     &  57.8    &  71.4     & - \\
    G-RMI \citep{papandreou2017towards} & ResNet-101   &42.6M & 57.0  &64.9   & 85.5      &  71.3     &   62.3    &   70.0    &   69.7 \\
    G-RMI* \citep{papandreou2017towards} & ResNet-101  &42.6M & 57.0   &68.5    &87.1      &75.5       &  65.8     &   73.3     & 73.3 \\
    RMPE \citep{fang2017rmpe} & Hourglass & 28.1M & 36.7 &72.3      &89.2     &79.1       &68.0       &78.6         &- \\
    CPN \citep{chen2018cascaded} & ResNet-Inception & - & -   &72.1      &91.4      &80.0     &68.7        &77.2         &78.5 \\
    CPN+ \citep{chen2018cascaded} & ResNet-Inception & - & -   &73.0     & 91.7    &80.9       &69.5       &78.1         &79.0 \\
    Baseline \citep{xiao2018simple} & ResNet-152 & 68.6M & 35.9 & 73.7     &  91.9     &  81.1     &  70.3     &  80.0     & 79.0 \\
    Baseline+* \citep{xiao2018simple} & ResNet-152 & - & - & 76.5     &  92.4     &  84.0     &  73.0     & 82.7     & 81.5 \\
    HRNet \citep{sun2019deep} & HRNet-w48 & 63.6M & 35.4 & 75.5     &  92.5     &  83.3     &  71.9     &  81.5    & 80.5 \\
    HRNet* \citep{sun2019deep} & HRNet-w48 &  63.6M & 35.4 & 77.0   &  92.7     &  84.5     &  73.4     &  83.1    & 82.0 \\
    MSPN \citep{li2019rethinking} & 4xResNet-50 & 71.9M & 58.7 &76.1  &93.4      &83.8       &72.3       &81.5       &81.6 \\
    MSPN* \citep{li2019rethinking} & 4xResNet-50 & 71.9M & 58.7 &77.1  &93.8      &84.6       &73.4       &82.3       &82.3 \\
    MSPN+* \citep{li2019rethinking}$\dagger$ & 4xResNet-50 & - & - &78.1  &{94.1}       &{85.9}       &74.5       &{83.3}       &{83.1} \\
    DARK \citep{zhang2020distribution} & HRNet-w48 & 63.6M & 32.9 & 76.2     &  92.5     &  83.6     &  72.5     &  82.4    & 81.1 \\
    DARK* \citep{zhang2020distribution} & HRNet-w48 &  63.6M & 32.9 & 77.4 &  92.6     &  84.6     &  73.6     &  83.7    & 82.3 \\
    RSN \citep{cai2020learning} & 4xRSN-50 & 111.8M & 65.9 & 78.6  & 94.3      &86.6     &75.5     &83.3       &83.8 \\
    RSN+ \citep{cai2020learning}$\ddagger$ & 4xRSN-50 & - & - &\textbf{79.2}  &\textbf{94.4}       &\textbf{87.1}       &\textbf{76.1}       &\textbf{83.8}       &\textbf{84.1} \\
    \midrule
    \midrule
    \textbf{CCM} & ResNet-152 & 63.5M & 40.1 &  75.8   &  92.7     &  83.4     &  71.8     &  81.5   & 80.9 \\
    \textbf{CCM} & HRNet-w48 & 63.7M & 36.6 & 76.6    &  92.8     &  84.1     &  72.6     &  82.4    & 81.7 \\
    \textbf{CCM*} & ResNet-152 & 63.5M & 40.1 & 77.3    &  93.0     &  84.8     &  73.3     &  83.1   & 82.3 \\
    \textbf{CCM*}  & HRNet-w48 & 63.7M & 36.6 & 78.0   &  93.4     &  85.1     &  74.0    &  83.6    & 83.0 \\
    \textbf{CCM+*} & HRNet-w48& - & - & \textbf{78.9}    &  \textbf{93.8}     &  \textbf{86.0}     &  \textbf{75.0}     &  \textbf{84.5}    & \textbf{83.6} \\
    \bottomrule
    \end{tabular}%
  \label{tab:SOTA_test}%
\end{table*}%

Our large model based on ResNet-152 with input size $384\times288$ achieved a gain of 2.4 AP over the simple baseline model \citep{xiao2018simple} and a gain of 0.4 AP over the recent HRNet-w48 model \citep{sun2019deep}. For example, CCM outperformed HRNet-w48 by a margin of 0.4 AP and 0.5 AP at the threshold 0.5 and 0.75. However, CCM was inferior to HRNet-w48 for large person instances, i.e., a drop of 0.2 AP. One possible explanation is that HRNet-w48 learned a high-resolution and discriminative feature representation by integrating the features from different scales. We attached a CM to HRNet-w48 and used the sub-pixel refinement techniques for postprocessing. It further improved the detection accuracy of HRNet-w48 from 76.3 AP to 77.5 AP. Besides, the result of large person instances was improved by 0.6 AP. Our CCM model achieved the best performance among all other models based on comparable backbone networks. Besides, both models have nearly the same parameters and GFLOPs as the baseline models since we decreased the number of filters in the CM for ResNet-152 and only used one CM for HRNet-w48. These results demonstrate the effectiveness of the proposed CCM, training strategies, and sub-pixel refinement techniques. 

The results of CCM and SOTA methods on the COCO test-dev set are summarized in Table~\ref{tab:SOTA_test}. The CCM using ResNet-152 as the backbone network trained on the COCO dataset outperformed the baseline model \citep{xiao2018simple} by 2.1 AP. It also outperformed the recent HRNet-w48 model \citep{sun2019deep} by 0.3 AP. Using HRNet-w48 as the backbone network, CCM outperformed the vanilla HRNet-w48 model, MSPN \citep{li2019rethinking}, and DARK \citep{zhang2020distribution} by a margin of 1.1 AP, 0.5 AP, and 0.4 AP, respectively. It was even better than the ensemble simple baseline models trained with the external AI Challenger dataset. After joint-training with this external dataset, CCM based on ResNet-152 outperformed both the simple baseline method and HRNet-w48. Besides, replacing the backbone network from ResNet-152 to HRNet-w48, the performance was further improved by 0.7 AP. It was better than the recently proposed method DARK* \citep{zhang2020distribution} by a margin of 0.6 AP using the same person detection results and comparable with the ensemble models MSPN+* \citep{li2019rethinking} (the champion of the 2018 COCO Keypoint Challenge), i.e., 78.0 v.s. 78.1. Generally, the external dataset brought about 1.5 AP improvement, which mainly arose from the AP at the larger threshold, demonstrating that training on more diverse poses helped the model to learn discriminative features and improve the location accuracy. Our final ensemble models brought another 0.9 AP and set a new state-of-the-art on this benchmark, i.e., 78.9 AP. It was comparable with RSN+ \citep{cai2020learning} (the champion of the 2019 COCO Keypoint Challenge), which used a better person detector and had 59.8 AP for the person category on COCO val2017.

\begin{figure*}[htbp]
\begin{center}
   \includegraphics[width=1\linewidth]{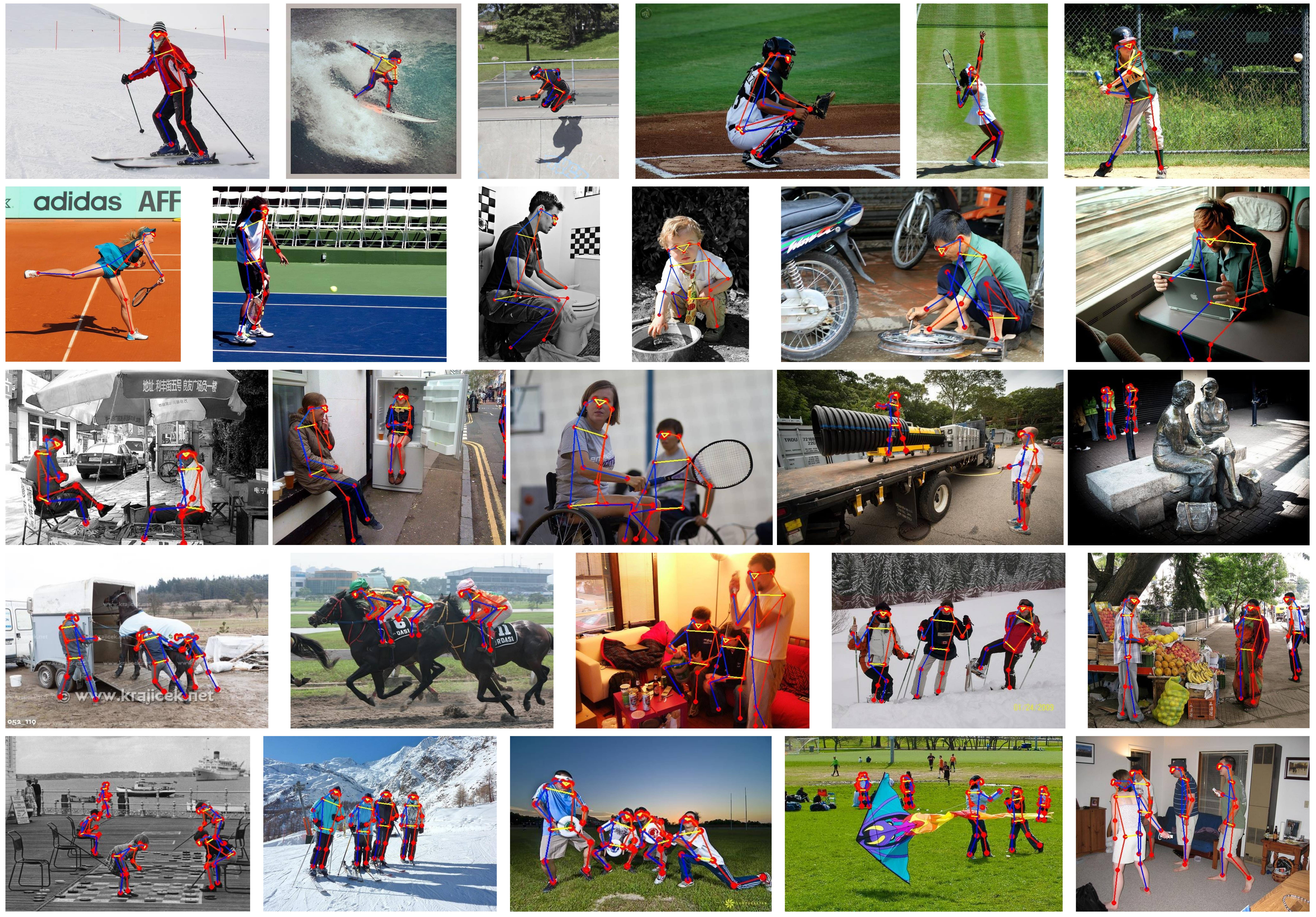}
\end{center}
   \caption{Some visual examples of the keypoint detection results on the COCO minival set \citep{lin2014microsoft}.}
\label{fig:visualDet}
\end{figure*}

\begin{figure*}[htbp]
\begin{center}
   \includegraphics[width=1\linewidth]{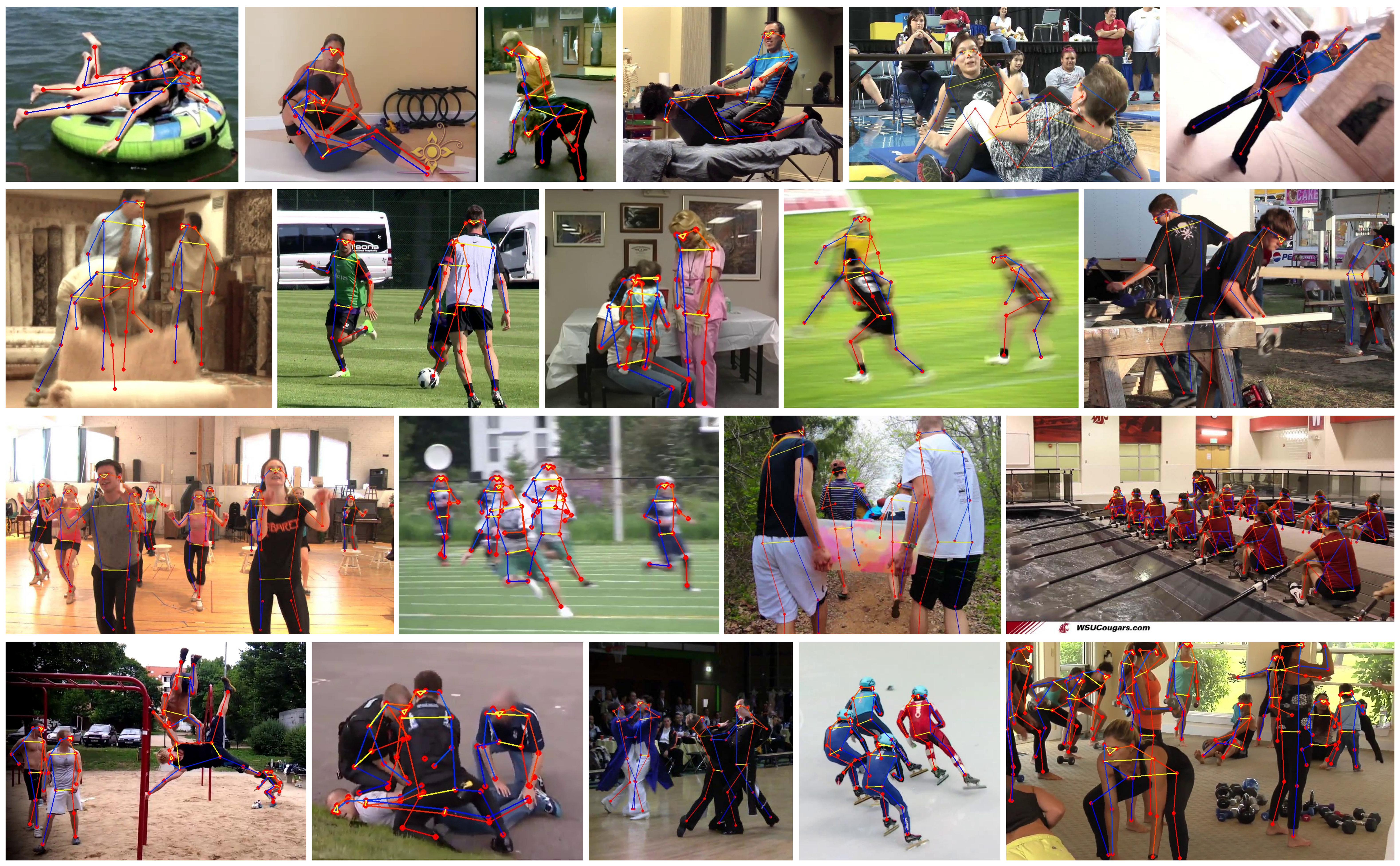}
\end{center}
   \caption{Some visual examples of the keypoint detection results on the PoseTrack validation set \citep{Andriluka2018PoseTrack}.}
\label{fig:visualDet_posetrack}
\end{figure*}

\begin{figure*}[t]
\begin{center}
   \includegraphics[width=1\linewidth]{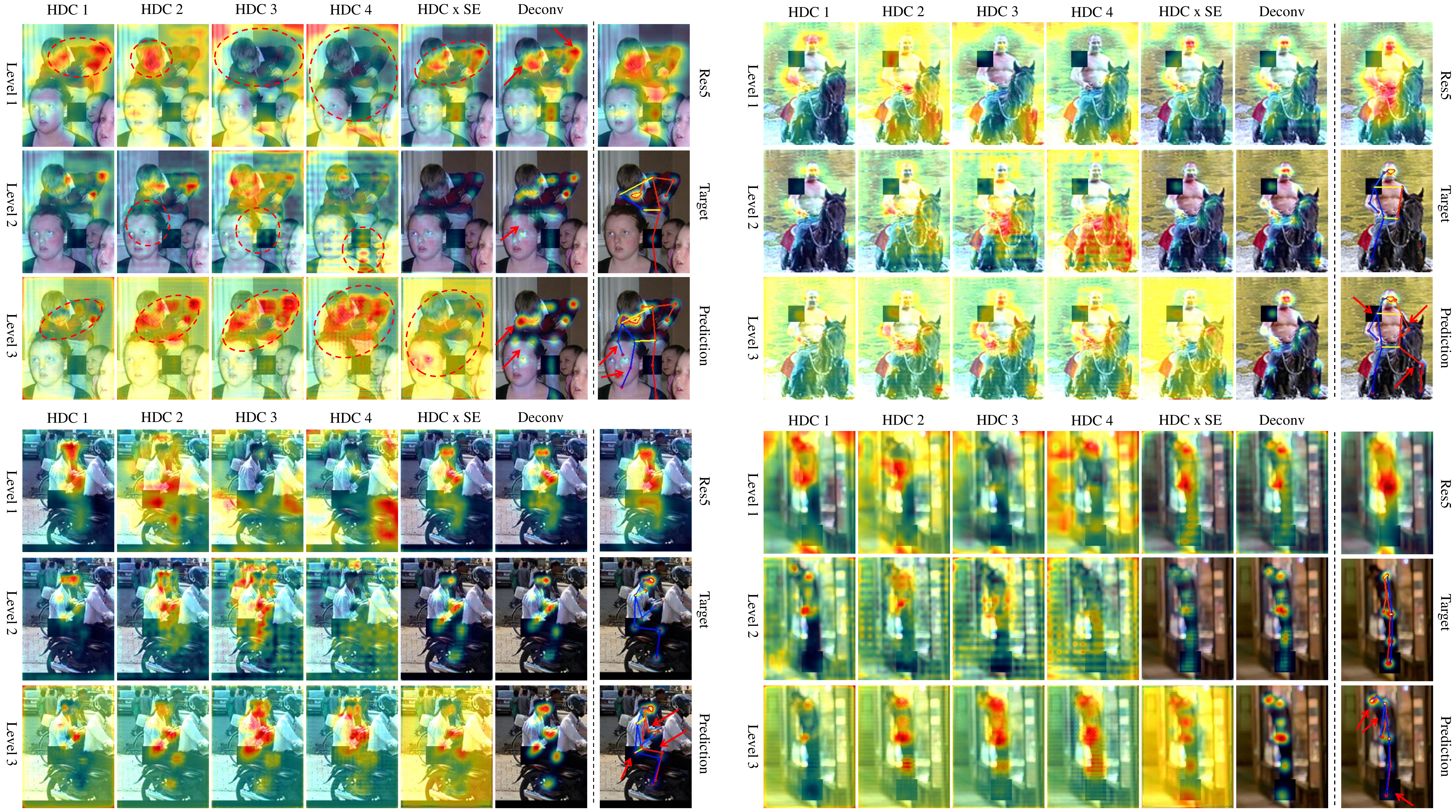}
\end{center}
   \caption{Visualization of the feature maps from the CMs in CCM based on ResNet-50. ``HDC k'' stands for the feature maps from the $k${th} dilated convolutional layer in the hybrid-dilated convolutional branches. ``HDC x SE'' stands for the first term in Eq.~\eqref{eq:feat_cam}. ``Deconv'' stands for the outputted feature maps from the CMs, i.e., the left side of Eq.~\eqref{eq:feat_cam}.}
\label{fig:context}
\end{figure*}

\begin{figure}[!ht]
\begin{center}
   \includegraphics[width=1\linewidth]{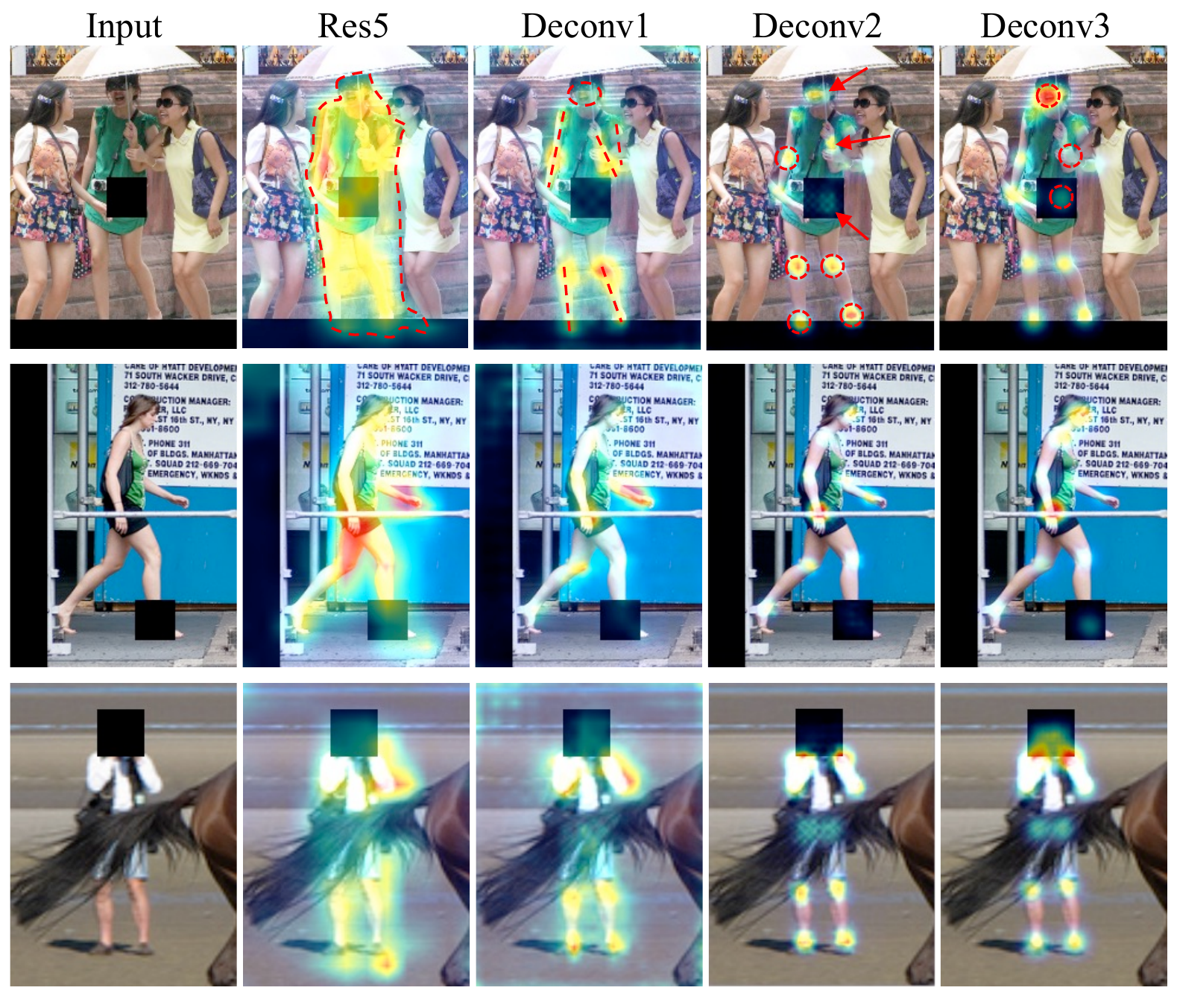}
\end{center}
   \caption{Visualization of the feature maps learned by CCM based on ResNet-50 at different stages. ``Deconv'' stands for the outputted feature maps from the CMs, i.e., the left side of Eq.~\eqref{eq:feat_cam}. One or several of keypoints of each person is manually occluded by a mask.}
\label{fig:feature_vis}
\end{figure}

\begin{figure}[!ht]
\begin{center}
   \includegraphics[width=1\linewidth]{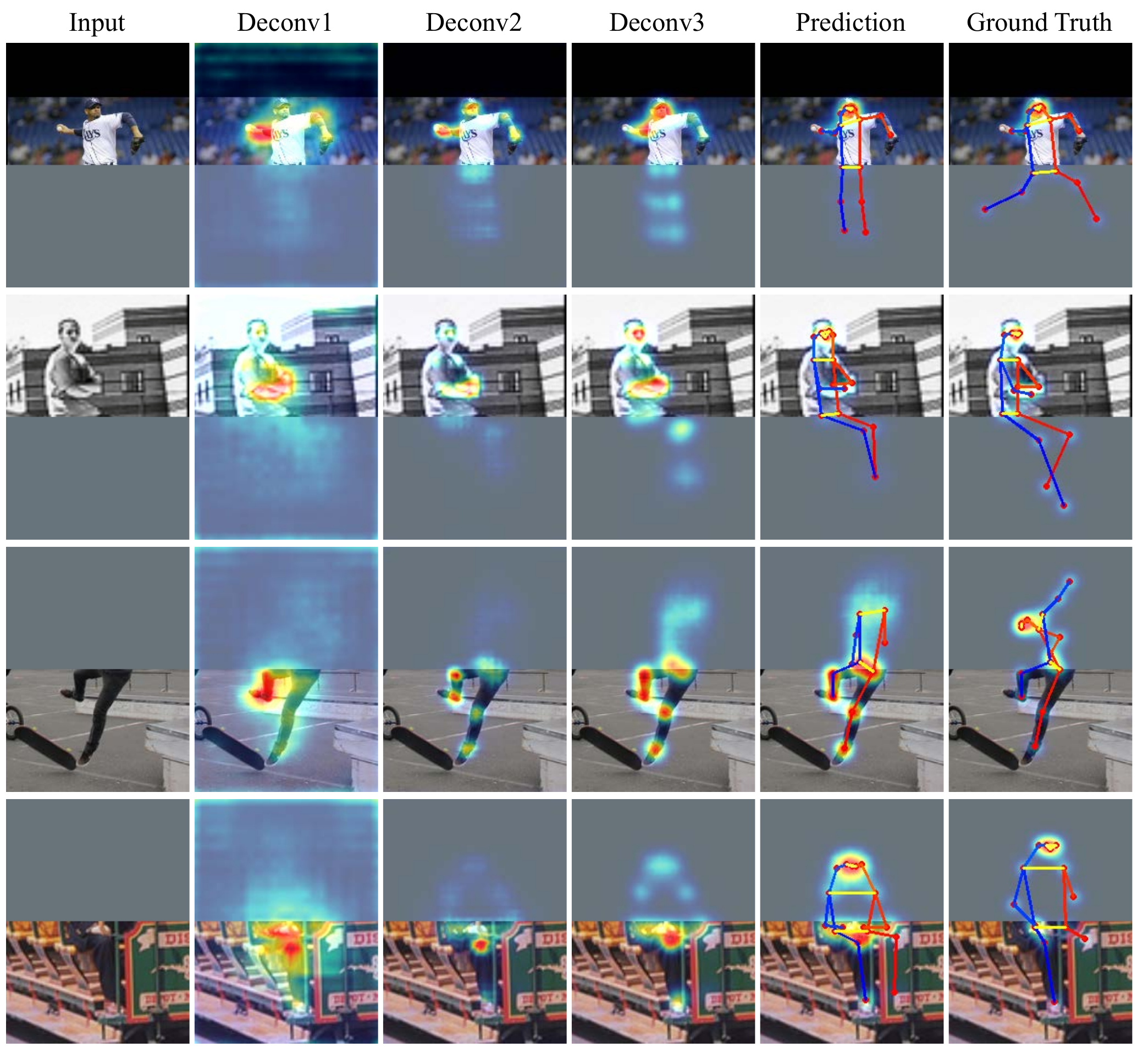}
\end{center}
   \caption{Visualization of the feature maps learned by CCM based on ResNet-50 at different stages. Either the upper or lower body of each person is manually masked out.}
\label{fig:feature_vis_mask}
\end{figure}

\textbf{Remarks}: 1) Our CCM model consistently outperforms the baseline models using small, medium, and large backbone networks, but the gain becomes smaller with the increasing of the model capacity, i.e., MobileNet $\approx$ ShuffleNet $\approx$ ResNet-50 $\approx$ ResNeXt-50 $\ge$ HRNet-w32 $\approx$ ResNet-152 $\ge$ HRNet-w48. It is reasonable because ``bigger'' backbone networks themselves have stronger representation capacity, thereby the impact of CM decreases accordingly; and 2) our CCM model benefits from the effective CM modules to model the context information, the efficient training strategies to learn discriminative features, and the sub-pixel refinement techniques to locate keypoints accurately, in a collaborative and complementary manner.  

\subsection{Subjective visual inspection and discussion}
\label{subsec:subjective}
We presented some visual examples of the keypoint detection results by using the CCM model based on HRNet-w48 on the COCO minival set and PoseTrack validation set in Figure~\ref{fig:visualDet} and Figure~\ref{fig:visualDet_posetrack}, respectively. As can be seen from, our model could handle various poses. Besides, it also successfully inferred the occluded keypoints (self-occluded or occluded by other objects). In the bottom two rows, we presented the detection results on multiple person instances within each image, which were also promising. Our model could handle small instances, blurry ones, low-light images as well as various occlusions. To see how CCM achieved the performance, we conducted an experiment to visually inspect the learned features by the network. 

We overlaid the output feature maps from the CMs in CCM on the input images to inspect what has been learned by the network. Figure~\ref{fig:context} shows the feature maps learned by the CMs at different levels. ``HDC k'' stands for the feature maps from the $k$th dilated convolutional layer in the hybrid-dilated convolutional branches. ``HDC x SE'' stands for the first term in Eq.~\eqref{eq:feat_cam}. 'Deconv' stands for the outputted feature maps from the CMs, i.e., the left side of Eq.~\eqref{eq:feat_cam}. ``Level k'' stands for the index of CM in CCM. The keypoints belonging to the left (right) body were connected by red (blue) lines. The left-right symmetric keypoints were connected by yellow lines. The predicted invisible keypoints of occluded body parts were indicated by red arrows.

As can be seen, HDC paid more attention to the context with the increase of the dilation rate. CCM learned to identify easy keypoints and inferred hard keypoints progressively through the cascaded CMs. Please check the arrows and ellipses. Moreover, 1) CCM probably learned the body configuration by inferring the invisible keypoints and potential poses conditioned on the visible keypoints as shown in the first row; 2) CCM also learned the body symmetry, e.g., there were two legs and arms in the human body, as shown in the left part of the second row; 3) CCM could also handle blurry images and infer the head and right arms as shown in the right part of the second row.

To illustrate the effectiveness of CCM for handling occlusions, we manually added masks on some body parts, \emph{e.g.}, the head, hip, and ankle as supplements to the existing occlusions, as shown in Figure~\ref{fig:feature_vis}. CCM first recognized and located the human bodies using the encoder (Res5), then detected different body parts (Deconv1) to help identify some distinct keypoints (Deconv2). In the final stage (Deconv3), CCM predicted the difficult occluded keypoints using the context information of identified body parts and keypoints. To further investigate the effectiveness of CCM for handling severe occlusions, we manually masked out either the upper or lower body of each person as shown in the first column of Figure~\ref{fig:feature_vis}. As can be seen, although the predicted human poses were different from the ground truth annotations at the masked regions, CCM showed the ability to learn human body configuration and infer reasonable invisible keypoints. Moreover, the results also confirm that CCM detected those distinct keypoints in the earlier stage while predicting the difficult occluded ones in the later stage.

\textbf{Remarks}: 1) Empirically, CCM's detection process follows a ``Localization $\to$ Componentization $\to$ Identification $\to$ Prediction'' routine, similar to the procedure that humans detect human keypoints in occluded settings (Section~\ref{subsec:motivation}); and 2) CCM probably has learned the body configuration such as symmetric body parts and reasonable distances between adjacent keypoints, evidenced by visual examples.

\begin{table*}[htbp]
  \centering
  \small
  \caption{Comparison between CCM-I and CCM-V on the COCO minival set. Please refer to Section~\ref{subsec:insight}.}
    \begin{tabular}{llllllllllll}
    \toprule
      Model  & Backbone   & $AP$    & $AP^{@.5}$  & $AP^{@.75}$ & $AP^M$ & $AP^L$ & $AR$  & $AR^{@.5}$  & $AR^{@.75}$ & $AR^M$ & $AR^L$ \\
    \midrule
    CCM-I & ResNet-50 & 70.4  & 89.1  & 76.7  & 66.0  & 77.0  & 77.1  & 93.4  & 82.9  & 72.2  & 83.8 \\
    CCM-V & ResNet-50 & 73.7  & 90.0  & 80.9  & 69.5  & 79.9  & 79.2  & 93.5  & 85.5  & 74.7  & 85.5 \\
    CCM & ResNet-50 & 73.8  & 90.2  & 80.9  & 69.6  & 80.1  & 79.3  & 93.7  & 85.6  & 74.8  & 85.7 \\
    \bottomrule
    \end{tabular}%
  \label{tab:main_results}%
\end{table*}%

\begin{table*}[htbp]
  \centering
  \small
  \caption{Comparison between CCM-I and CCM-V on the visible and invisible keypoints in the COCO minival set.}
    \begin{tabular}{lllllllllllll}
    \toprule
      Model  & Backbone  &   Kpt. Type    & $AP$    & $AP^{@.5}$  & $AP^{@.75}$ & $AP^M$ & $AP^L$ & $AR$  & $AR^{@.5}$  & $AR^{@.75}$ & $AR^M$ & $AR^L$ \\
    \midrule
    \multirow{2}[0]{*}{CCM-I} & \multirow{2}[0]{*}{ResNet-50} & Visible   & 79.3  & 94.6  & 86.7  & 77.4  & 82.2  & 82.4  & 95.5  & 88.7  & 80.2  & 85.5 \\
        \cmidrule{3-13}
        &  & Invisible & 40.2  & 64.2  & 39.5  & 40.2  & 42.3  & 47.8  & 69.8  & 48.0    & 45.1  & 53.2 \\
    \midrule
    \midrule
    \multirow{4}[0]{*}{CCM-V} & \multirow{4}[0]{*}{ResNet-50}& Visible   & 79.4  & 94.7  & 87.0    & 76.9  & 82.4  & 82.0    & 95.3  & 88.1  & 79.5  & 85.4 \\
         & & $\pm \Delta$  & 0.1   & 0.1   & 0.3   & -0.5  & 0.2   & -0.4  & -0.2  & -0.6  & -0.7  & -0.1 \\
          \cmidrule{3-13}
        &  & Invisible & 55.0    & 79.4  & 57.1  & 54.7  & 57.8  & 61.4  & 82.1  & 63.9  & 58.8  & 66.9 \\
         & & $\pm \Delta$  & 14.8  & 15.2  & 17.6  & 14.5  & 15.5  & 13.6  & 12.3  & 15.9  & 13.7  & 13.7 \\
    \bottomrule
    \end{tabular}%
  \label{tab:vis_invis}%
\end{table*}%

\begin{table*}[htbp]
  \centering
  \small
  \caption{Comparison between CCM-I and CCM-V on different types of instances w.r.t. the numbers of annotated keypoints.}
    %\begin{tabular}{cccccccccccccccccccc}
    \begin{tabular}{p{1cm}p{0.6cm}p{0.4cm}p{0.6cm}p{0.4cm}p{0.4cm}p{0.4cm}p{0.4cm}p{0.4cm}p{0.4cm}p{0.4cm}p{0.4cm}p{0.4cm}p{0.4cm}p{0.4cm}p{0.4cm}p{0.4cm}p{0.4cm}p{0.4cm}}
    \toprule
    & \multicolumn{17}{c}{Number of annotated keypoints per instance} &\\
    \cmidrule{3-19}
      Model  & Metric & 1     & 2     & 3     & 4     & 5     & 6     & 7     & 8     & 9     & 10    & 11    & 12    & 13    & 14    & 15    & 16    & 17 \\
    \midrule
    CCM-I & AP & 32.7  & 25.0    & 33.4  & 38.9  & 48.1  & 52.5  & 60.2  & 63.4  & 67.1  & 68.0    & 74.2  & 77.8  & 79.0    & 81.6  & 82.6  & 88.8  & 90.9 \\
    \midrule
    \midrule
    \multirow{2}[0]{*}{CCM-V} & AP& 33.9  & 23.9  & 37.6  & 43.6  & 48.2  & 58.1  & 64.2  & 70.0    & 72.6  & 72.8  & 77.9  & 81.5  & 81.7  & 83.4  & 84.1  & 89.3  & 91.2 \\
          &$\pm \Delta$ & 1.2   & -1.1  & \textbf{4.2} & \textbf{4.7} & 0.1   & \textbf{5.6} & \textbf{4.0} & \textbf{6.6} & \textbf{5.5} & \textbf{4.8} & 3.7   & 3.7   & 2.7   & 1.8   & 1.5   & 0.5   & 0.3 \\
    \bottomrule
    \end{tabular}%
  \label{tab:ap_kptsNum}%
\end{table*}%

\subsection{Empirical Studies on Annotations}
\label{subsec:insight}
As we know that occluded keypoints are more difficult to be detected than visible ones, we are wondering whether the invisible keypoint annotations have the same impact on the model or not, compared with the same amount of visible keypoint annotations? To this end, we conducted an experiment to gain some insight into the keypoint annotations.

We constructed two training sets based on the COCO training set. The first one was COCO-I which was obtained by removing all the annotations of invisible keypoints in the original training set. The second one was COCO-V which was obtained by removing the same amount of visible keypoint annotations randomly. The resulting keypoints without annotations were treated as unlabeled. Then, we trained the proposed CCM using the ResNet-50 as the backbone encoder on COCO-I and COCO-V. They were denoted as ``CCM-I'' and ``CCM-V'', respectively. Their results on the COCO minival set are summarized in the first two rows of Table~\ref{tab:main_results}. As a reference, we also listed the model trained on the original COCO training set in the bottom row.

As can be seen, the scores of CCM-V dropped marginally compared with CCM. However, the scores of CCM-I dropped significantly by a large margin compared with CCM-V and CCM. These results confirm that \emph{{annotating invisible keypoints matters}}, \emph{i.e.}. The invisible keypoint annotations contributed more to the model than the same amount of visible ones. Since it is more difficult to detect invisible keypoints, the invisible keypoint annotations provide stronger supervisory signals to the model than the visible ones did. To infer an invisible keypoint with such supervision, it probably learned useful features from the context since the keypoint itself was invisible. In this way, the model could learn the knowledge of body configuration implicitly, e.g., as a form of discriminative feature for each category of keypoints.

To further analyze the impact of invisible keypoint annotations, we calculated the indexes on visible and invisible keypoints, respectively. We removed all the invisible keypoint annotations from the COCO minival set. The resulting minival set was used to evaluate the model's performance on the visible keypoints. Similarly, we also removed all the visible keypoint annotations from the COCO minival set to evaluate the model's performance on the invisible keypoints. We used the ground truth bounding boxes as the person detection results. The results are listed in Table~\ref{tab:vis_invis}. $\pm \Delta$ denoted the gain of CCM-V over CCM-I.

Unsurprisingly, CCM-V achieved better results on invisible keypoints compared with CCM-I, demonstrating that the gains of CCM-V over CCM-I in Table~\ref{tab:main_results} mainly arose from the invisible ones. Note that although CCM-I was trained without using the annotations of invisible keypoints, it still obtained 40.2 AP on them, demonstrating that it had a bit of generalization on predicting the invisible keypoints. Since there were distinct appearance differences between visible keypoints and invisible ones of the same category and the model did not get any supervisory signal from the invisible keypoints, it implies that CCM probably learned useful features from contextual keypoints to infer the invisible ones. Using the invisible annotations, CCM achieved better performance by exploiting its representation capacity.

We also reported APs of the two models on instances with different numbers of annotated keypoints from the COCO minival set. The results are shown in Table~\ref{tab:ap_kptsNum}. As can be seen, the gains mainly arose from the occluded instances, for example, instances with less than 10 annotated keypoints. It implies that the invisible keypoint annotations helped the model to learn the body configuration for inferring the invisible keypoints and handling occlusions. Some visual results were presented in Figure~\ref{fig:context} and Figure~\ref{fig:feature_vis}.

Although annotating invisible keypoints is more difficult than visible ones, the above results confirm that the invisible keypoint annotations are more valuable. Consequently, we could improve our model by 1) annotating more occluded keypoints to train a better model; 2) exploiting the active learning strategy to identify hard keypoints that should be annotated to continuously improve the model; 3) developing effective multi-view learning algorithms to utilize the complementary information between different views of data.

\textbf{Remarks}: 1) The invisible keypoint annotations have a larger impact on the model than the same amount of visible ones; and 2) even without the invisible annotations, the proposed CCM model is still able to generalize to the invisible keypoints.

\section{Conclusion}
In this paper, we address the human keypoint detection problem by devising a new cascaded context mixer-based neural network (CCM), which has a strong representative capacity to simultaneously model the spatial and channel context information. We also propose three efficient training strategies including a hard-negative person detection mining strategy to migrate the mismatch between training and testing, a joint-training strategy to use abundant unlabeled samples by knowledge distilling, and a joint-training strategy to exploit external data with heterogeneous labels. They collaboratively enable CCM to learn discriminative features from abundant and diverse poses. Besides, we present four postprocessing techniques to refine predictions at the sub-pixel level accuracy. These complementary techniques are carried out sequentially during the inference phase for unique and explicit purposes, which further improve the detection accuracy. Our CCM model consistently outperforms public state-of-the-art models with various backbone networks by a large margin. We empirically show that CCM’s detection process is similar to humans in occluded settings and probably learns human body configuration. Moreover, we also identify that invisible keypoint annotations have a larger impact on the model than the same amount of visible ones and present some promising research topics.

%\begin{acknowledgements}
%\end{acknowledgements}



\bibliographystyle{spbasic}      % basic style, author-year citations
\bibliography{2020IJCV_humanPose}   % name your BibTeX data base

%\begin{thebibliography}{}
%%
%% and use \bibitem to create references. Consult the Instructions
%% for authors for reference list style.
%\end{thebibliography}

\end{document}